\documentclass[10pt,twocolumn,letterpaper]{article}
\usepackage[T1]{fontenc}
\usepackage[utf8]{inputenc}
\usepackage{times}
\usepackage[letterpaper,textwidth=6.875in,textheight=8.875in,top=0.85in,columnsep=0.3125in]{geometry}
\usepackage{amsmath,amssymb,graphicx,booktabs,xspace}
\usepackage[dvipsnames,table]{xcolor}
\usepackage{tabularx,multirow}
\usepackage{float,placeins,dblfloatfix}
\usepackage{microtype,etoolbox}
\usepackage[numbers,sort&compress]{natbib}
\usepackage[format=plain,labelformat=simple,labelsep=period,font=small,compatibility=false]{caption}
\usepackage[font=footnotesize,skip=3pt,subrefformat=parens]{subcaption}
\usepackage[hyphens]{url}
\usepackage[shortlabels,inline]{enumitem}
\definecolor{linkblue}{RGB}{40,85,130}
\usepackage[breaklinks,colorlinks,allcolors=linkblue]{hyperref}
\usepackage[capitalize]{cleveref}
\hypersetup{
  pdftitle={ReCue: Hidden-State Recycling for Vision-Language Speculative Decoding},
  pdfauthor={Ziyang Ma, Zihong Zhang, Lefei Zhang, Baoyuan Qi, Zuchao Li},
  pdfsubject={Vision-language speculative decoding},
  pdfkeywords={vision-language models, speculative decoding, hidden-state recycling}
}

\makeatletter
\renewcommand{\section}{\@startsection{section}{1}{\z@}%
  {-2.2ex plus -0.5ex minus -0.2ex}{1.0ex plus 0.2ex}{\normalfont\large\bfseries}}
\renewcommand{\subsection}{\@startsection{subsection}{2}{\z@}%
  {-1.8ex plus -0.4ex minus -0.2ex}{0.8ex plus 0.2ex}{\normalfont\normalsize\bfseries}}
\renewcommand{\subsubsection}{\@startsection{subsubsection}{3}{\z@}%
  {-1.5ex plus -0.4ex minus -0.2ex}{0.6ex plus 0.2ex}{\normalfont\normalsize\bfseries}}
\renewcommand{\paragraph}{\@startsection{paragraph}{4}{\z@}%
  {1.5ex plus 0.4ex minus 0.2ex}{-1em}{\normalfont\normalsize\bfseries}}
\makeatother
\newcolumntype{Y}[1]{>{\hsize=#1\hsize\linewidth=\hsize\centering\arraybackslash}X}
\definecolor{recueTableHighlight}{RGB}{234,242,248}
\newcommand{\recueMetricPair}[2]{%
    \makebox[18pt][r]{#1}\hspace{4pt}\makebox[28pt][r]{#2}}
\newcommand{\recueMetricHeader}{%
    \recueMetricPair{\scriptsize MAT}{\scriptsize Speedup}}
\newcommand{\recueMethodGroup}[2]{%
    \multicolumn{#1}{l}{\footnotesize\itshape #2}\\}
\newcommand{\recueMethodRule}{%
    \arrayrulecolor{black!40}\specialrule{0.25pt}{1pt}{0pt}\arrayrulecolor{black}}

\title{ReCue: Hidden-State Recycling for\\Vision-Language Speculative Decoding}
\author{\textnormal{Ziyang Ma\textsuperscript{1,$\dagger$}},
  \textnormal{Zihong Zhang\textsuperscript{2,$\dagger$}},
  \textnormal{Lefei Zhang\textsuperscript{1,$*$}},
  \textnormal{Baoyuan Qi\textsuperscript{3}},
  \textnormal{Zuchao Li\textsuperscript{2,$*$}} \\
  {\small\textsuperscript{1}School of Computer Science, Wuhan University, Wuhan, China} \\
  {\small\textsuperscript{2}School of Artificial Intelligence, Wuhan University, Wuhan, China} \\
  {\small\textsuperscript{3}Xiaomi, Beijing, China} \\
  {\footnotesize\texttt{\{maziyang, zhangzihong, zhanglefei, zcli-charlie\}@whu.edu.cn}} \\
  {\footnotesize\texttt{qibaoyuan@xiaomi.com}}}
\date{}

\makeatletter
\renewcommand{\@maketitle}{%
    \begin{center}
        {\fontsize{19}{22}\selectfont\bfseries\@title\par}
        \vspace{9pt}
        {\normalsize\begin{tabular}[t]{c}\@author\end{tabular}\par}
        \vspace{7pt}
    \end{center}
    \begin{minipage}{\textwidth}
        \centering
        \includegraphics[width=0.90\textwidth]{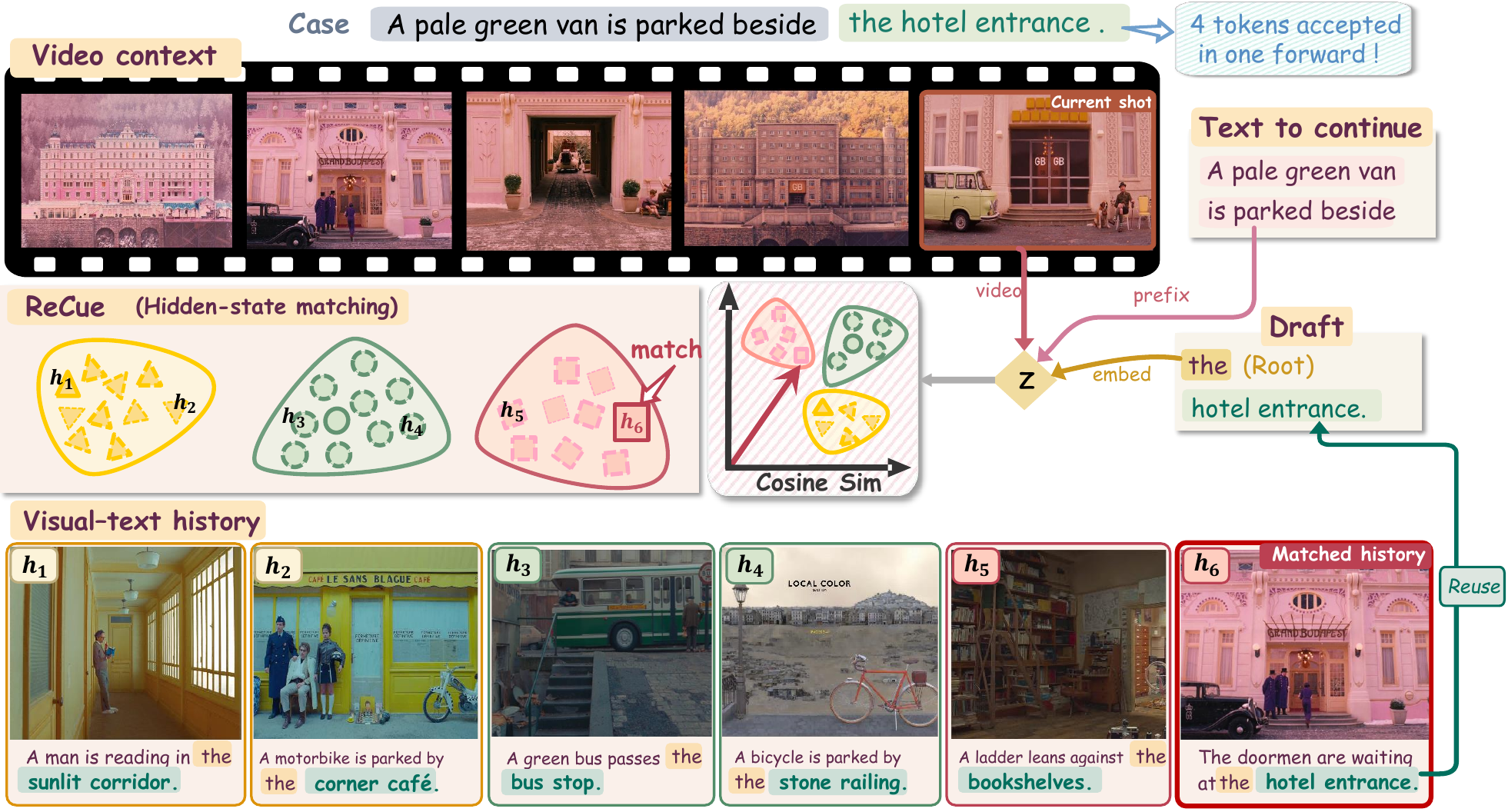}
        \captionsetup{type=figure,skip=1pt}
        \captionof{figure}{Hidden-state recycling in ReCue. Visual inputs sharing the root token ``the'' can require different successors. The current visually conditioned state retrieves and ranks cached candidates for parallel verification. The video and accepted continuation are both illustrative.}
        \label{fig:recue_teaser}
    \end{minipage}\par\vspace{0.5\baselineskip}
}
\makeatother
\renewenvironment{abstract}{%
    \par\small\centerline{\bfseries Abstract}\vspace{6pt}\noindent\ignorespaces
}{\par\vspace{10pt}}

\begin{document}
\raggedbottom
\maketitle
\begingroup
\renewcommand{\thefootnote}{\fnsymbol{footnote}}
\footnotetext[2]{Equal contribution.}
\footnotetext[1]{Corresponding authors.}
\endgroup

\begin{abstract}
Autoregressive decoding remains a bottleneck for vision-language models. Speculative decoding reduces this cost through parallel verification, but text-matched reuse can preserve successor rankings that are incompatible with the current visual input. We introduce ReCue, which brings the target model’s visual awareness into cache-based draft-tree construction by recycling visually conditioned hidden states. A residual predictor or a same-root historical transition estimates the state after the draft root. The frozen target language head reranks a compact pool of current and retrieved candidates in this state, guiding continuation retrieval and tree expansion under a fixed node budget. Across three vision-language models, ReCue achieves pooled decode speedups of 2.68–3.05× on images and 2.37–2.57× on video over autoregressive decoding, with gains also demonstrated in SGLang.
Code: \url{https://github.com/whucs21Mzy/ReCue}.
\end{abstract}

\section{Introduction}
\label{sec:intro}

Vision-language models (VLMs) have demonstrated strong capabilities across diverse visual understanding and generation tasks. However, their autoregressive decoding remains a major source of inference latency, particularly when producing detailed descriptions and long-form responses. Speculative decoding (SD)~\cite{leviathan2023speculative} alleviates this bottleneck by verifying multiple drafted tokens in parallel. Among existing approaches, prediction recycling offers an attractive alternative to individual draft models by reusing target-model predictions and text continuations from previous computation. Methods such as Token Recycling~\cite{luo2025tokenrecycling}, RACER~\cite{zhang2026racer}, RECAP~\cite{zhu2026recap}, and LogitSpec~\cite{liu2026logitspec} exploit different forms of cached predictions, contextual retrieval, and lookahead to construct draft candidates with limited additional computation.

However, reusing predictions based on textual matches does not necessarily preserve their compatibility with the current visual context. Unlike text-only generation, the continuation of a VLM depends jointly on textual context and visual evidence. As illustrated in Fig.~\ref{fig:recue_teaser}, the same text token may admit substantially different successors depending on the visual content being described. Although cached predictions may have been obtained from visually conditioned model computations, their original rankings reflect the contexts in which they were generated. The prefix ``The car is'' can lead to different color descriptions across images, yet text-indexed caches may return identical successor orders. Reusing textual matches without adapting them to the current visual input can prioritize incompatible continuations. This limitation is particularly consequential for tree-based speculative decoding: under a fixed verification budget, selecting an incorrect early successor not only wastes a draft node but also prevents the acceptance of potentially correct continuations deeper along that branch.

This observation motivates a central question: Can we reuse the visual understanding already available in the target VLM to guide speculative draft construction? Existing vision-language speculative decoding methods have explored incorporating visual information into learned drafters, for example through compressed visual features or target-model hidden states~\cite{kang2025vispec, xie2026hivis}. We explore a complementary direction: rather than learning a separate autoregressive drafter, we leverage the target model's existing visually conditioned representations to adapt reusable predictions to the current multimodal context. Specifically, while the target's current hidden state determines the next token, predicting its successors requires a representation of the state after that token. Estimating this post-root state offers a lightweight way to recover visually conditioned successor preferences without an additional full target-model forward pass.

To this end, we propose \textbf{ReCue}, which recycles visually conditioned hidden states to incorporate visual awareness into draft construction.
ReCue estimates the post-root hidden state using either a compatible transition recycled from the request history or a small residual predictor. The target model's frozen language head then reranks a compact pool of current and cached successor candidates under this estimated state. Crucially, these visually conditioned candidate rankings do not merely determine the next drafted tokens: they also guide contextual continuation retrieval and fixed-budget draft-tree expansion. In this way, ReCue combines the efficiency of historical prediction reuse with the ability to adapt draft construction to the current visual input.

Across Qwen2.5-VL-7B, Qwen3-VL-8B, and LLaVA-NeXT Vicuna-7B, ReCue reaches pooled decode speedups of $2.68$--$3.05\times$ on images and $2.37$--$2.57\times$ on video over autoregressive decoding, while also demonstrating practical acceleration in SGLang.
Ablations validate the contributions of state recycling and contextual tree expansion.
Furthermore, controlled visual changes, region-level interventions, and image-swap analyses provide evidence that visual information influences successor ranking and the resulting draft acceptance.

\section{Related Work}
\label{sec:related_work}

\subsection{Speculative Decoding and Draft Reuse}
Speculative decoding reduces sequential target calls through parallel
verification~\cite{leviathan2023speculative,chen2023speculative}.
Medusa~\cite{cai2024medusa} uses multiple prediction heads;
EAGLE-2~\cite{li2024eagle2} adapts draft trees, and
EAGLE-3~\cite{li2025eagle3} fuses target features across layers.
DFlash~\cite{chen2026dflash} and DSpark~\cite{cheng2026dspark}
use diffusion and semi-autoregressive block drafting.

SSSD~\cite{marzollo2026sssd} combines CPU-based $n$-gram retrieval
with hardware-aware speculation.
Token Recycling~\cite{luo2025tokenrecycling} updates a token-adjacency cache.
RACER~\cite{zhang2026racer} combines contextual retrieval with copy-logit
expansion, reusing logits following the latest occurrence of the same
token to propose successors.
LogitSpec~\cite{liu2026logitspec} extends retrieval queries through
next-next-token speculation.

\subsection{Visual Conditioning and Video Acceleration}
For VLMs, effective draft construction can benefit from visual conditioning.
ViSpec~\cite{kang2025vispec} compresses visual tokens and conditions
text on a global image feature.
HiViS~\cite{xie2026hivis} passes visual information through target
hidden states at text positions.
DREAM~\cite{hu2025dream} uses target-feature cross-attention;
DREAM-S~\cite{liu2026dreams} searches draft architectures and
target interactions.
RECAP~\cite{zhu2026recap} fuses request-local and persistent prediction
memories by specificity and entropy.

Long visual sequences further increase drafting cost.
SpecVLM~\cite{ji2025specvlm} uses verifier-guided video-token pruning;
ParallelVLM~\cite{kong2026parallelvlm} overlaps prefill, drafting,
and verification.
HIPPO~\cite{lv2026hippo} preserves tokens using global, temporal,
and spatial signals and overlaps drafting with verification.
ReCue performs drafting by ranking current and retrieved
candidates through the frozen language head in a visually conditioned
post-root state.
This ranking guides retrieval and tree expansion under a fixed node budget.

\begin{figure*}[t]
    \centering
    \includegraphics[width=0.95\textwidth]{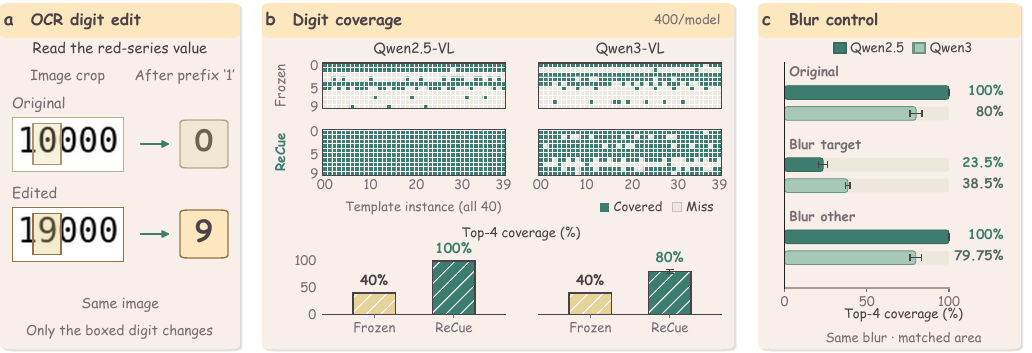}
    \caption{An OCR test of visual grounding.
    (a) Editing \texttt{10000} to \texttt{19000} changes the next digit after prefix \texttt{1} from \texttt{0} to \texttt{9}.
    (b) Top-4 digit coverage across all 400 cases per model: filled cells indicate hits; bars show overall coverage. Frozen reuses the first image's ranking.
    (c) Coverage of the original digit on the same cases after blurring the target or another digit with matched area and strength.
    Whiskers show 95\% template-bootstrap intervals. The prefix, source, and candidate pool are fixed within each template.}
    \label{fig:recue_visual_state}
\end{figure*}

\section{Preliminaries}
\label{sec:preliminaries}

\subsection{Speculative Decoding}
\label{sec:prelim_sd}

Speculative decoding~\cite{leviathan2023speculative,chen2023speculative} pairs inexpensive drafting with parallel target scoring. At round $t$, given visual input $I$ and the text prefix $P_t$ (the prompt and accepted output), a drafter samples a continuation $\widetilde y_{1:\gamma}$. The draft distribution $d_i$ and target distribution $p_i$ at position $i$ share the context $(I,P_t,\widetilde y_{<i})$. The target computes all $p_i$ in one pass, and verification processes the proposals in prefix order. For $\widetilde y_i\sim d_i$, the acceptance probability and the correction distribution are
\begin{equation}
    \begin{aligned}
        \Pr(\mathrm{accept}\mid\widetilde y_i)
            &=\min\!\left\{1,
                \frac{p_i(\widetilde y_i)}{d_i(\widetilde y_i)}\right\},\\
        p_i^{\mathrm{rej}}(v)
            &=\frac{[p_i(v)-d_i(v)]_+}
                    {\sum_u[p_i(u)-d_i(u)]_+},
    \end{aligned}
    \label{eq:speculative_acceptance}
\end{equation}
where $[z]_+=\max(z,0)$. At the first rejection, the verifier draws a correction from $p_i^{\mathrm{rej}}$ and discards the remaining draft tokens. If all $\gamma$ proposals pass, it draws a bonus token from the target after the accepted draft. The resulting sequence follows the target distribution.

\subsection{Greedy Tree Verification}
\label{sec:prelim_tree}

For greedy generation, verification follows the target argmax at each position. ReCue verifies alternative continuations in one pass by packing them into a tree with shared prefixes~\cite{cai2024medusa}, rooted at the known next token $r_t=\arg\max_v p(v\mid I,P_t)$. Let $n_0$ be the root node, $\mathcal K_t$ the cached prefix positions, and $\operatorname{Anc}^{+}(n)$ the positions of node $n$ and its ancestors. The attention mask and depth-based positions are
\begin{equation}
    \begin{aligned}
        M_t(n,u)&=
        \begin{cases}
            0, & u\in\mathcal K_t\cup\operatorname{Anc}^{+}(n),\\
            -\infty, & \text{otherwise},
        \end{cases}\\
        \operatorname{pos}(n)
            &=\operatorname{pos}(n_0)+\operatorname{depth}(n),
    \end{aligned}
    \label{eq:tree_attention}
\end{equation}
This gives each node the context it would receive if its branch were decoded separately. Write $B(n)$ for the token sequence from the root through $n$, $\tau(n)$ for its last token, and $\Pi_t$ for root-starting paths truncated at EOS. The target prediction after $n$ and the longest matching path are
\begin{equation}
    \begin{aligned}
        g_t(n)&=\arg\max_v p(v\mid I,P_t,B(n)),\\
        \pi_t^\star
            &=\operatorname*{arg\,max}_{\pi\in\Pi_t}|\pi|,\\
        \text{s.t.}\quad
        \tau(n_j)&=g_t(n_{j-1}),\quad j=1,\ldots,L,
    \end{aligned}
    \label{eq:greedy_tree_path}
\end{equation}
where $\pi=(n_0,\ldots,n_L)$. The accepted path includes the root and advances the prefix and KV cache. Its final node's prediction supplies the next root if generation continues. The same pass also produces hidden states and successor logits at every draft node for reuse in Sec.~\ref{sec:method}.

\begin{figure*}[t]
    \centering
    \includegraphics[width=\textwidth]{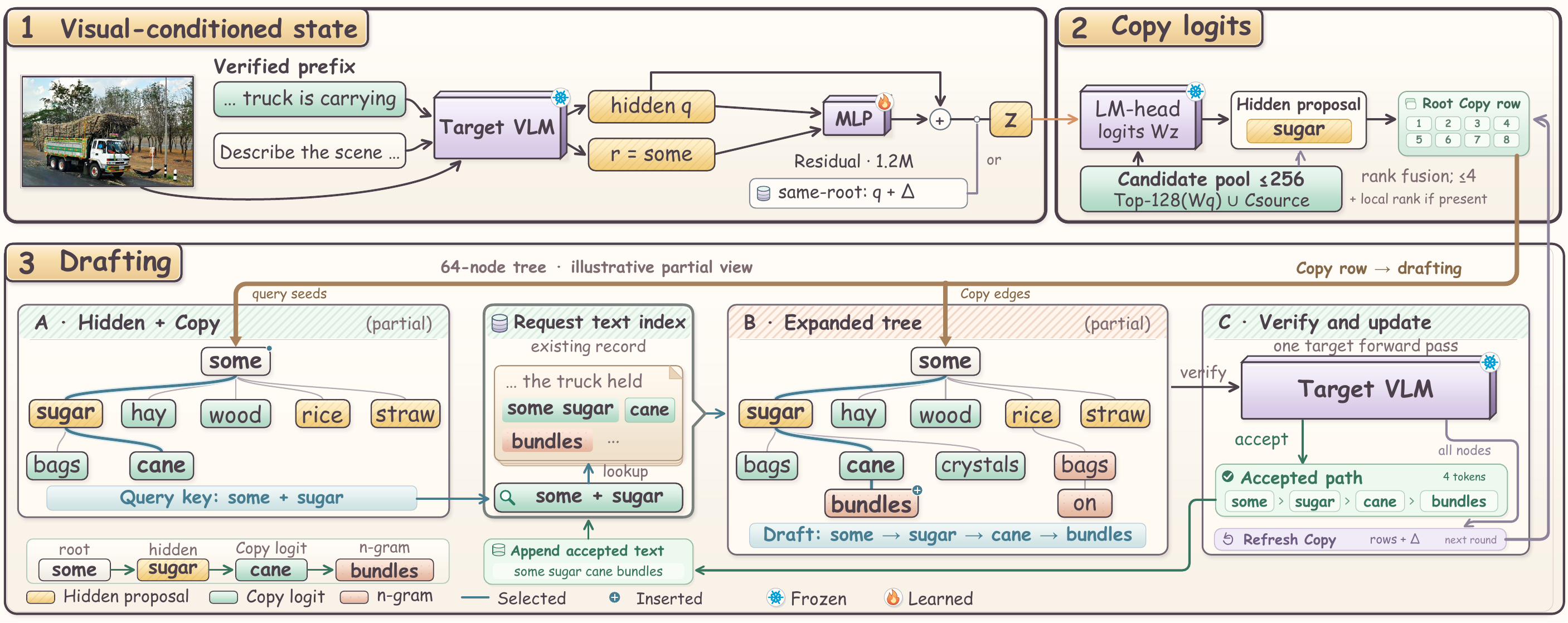}
    \caption{ReCue overview. A residual predictor or a recycled same-root transition constructs a proposal state from the current visually conditioned state. The frozen language head ranks a compact candidate pool, and selected successors seed a tree expanded by cached predictions and retrieved text. Target verification accepts a path and refreshes the prediction and transition memories.}
    \label{fig:recue_method}
\end{figure*}

\section{Method}
\label{sec:method}

\subsection{Overview}

\textbf{Motivation.} Prediction recycling accelerates speculative decoding by reusing cached target-model predictions and text continuations. However, textual matching alone cannot guarantee that retrieved predictions remain compatible with the current visual context. Given a visual input $I$, text prefix $P_t$, and target-selected root token $r_t$, drafting its successors requires approximating $p(v \mid I, P_t, r_t)$. Even with identical text and root tokens, different visual inputs may induce different successor distributions. As illustrated in Fig.~\ref{fig:recue_visual_state}, changing a single visual detail can alter the preferred successor without changing the textual context. Such mismatches can directly affect drafting efficiency. Under a fixed verification budget, incorrectly prioritized early successors occupy tree capacity and prevent potentially valid deeper continuations from being explored. Therefore, effective multimodal draft construction requires not only retrieving reusable predictions, but also adapting their priorities to the current visual input.
\textbf{Key insight.} The target VLM already encodes visual information in its hidden states. Rather than introducing a separate vision-aware drafter, we leverage these existing representations to recontextualize cached predictions. Specifically, while the current target state $q_t=h(I,P_t)$ predicts the root token $r_t$, evaluating its successors requires estimating the post-root state $h(I,P_t,r_t)$. This allows the target's frozen language head to reassess historical candidates under the current visual context without an additional full target-model forward pass.
\textbf{Framework.} Building on this insight, ReCue propagates visual information into draft construction through three stages (Fig.~\ref{fig:recue_method}). First, hidden-state recycling estimates the post-root representation using either a compatible historical transition or a lightweight residual predictor. Second, visually conditioned candidate reranking uses the frozen language head to reorder current and historical successor candidates. Third, contextual draft-tree construction uses these candidates to guide continuation retrieval and prediction-cache expansion within a fixed verification budget. The verified target predictions and hidden states are then recycled for subsequent decoding rounds.

\subsection{Hidden-State Recycling}

ReCue estimates the post-root state by applying a historical state transition to the current visually conditioned state $q_t$, rather than copying a historical hidden state. A request-local history $\mathcal H_t$ stores records of previously verified tokens $u_j$, their pre-token states $h_j^-$, state increments $\Delta_j=h_j^+-h_j^-$, and successor candidate lists $\mathcal C_j$. A fixed prediction bank $\mathcal B(r_t)$ provides additional sources indexed by the root token. We select a source $j_t$ from these memories, prioritizing exact-root matches and then state and token-embedding similarity.

A local transition is reused only when its token matches $r_t$; otherwise, a lightweight residual predictor estimates the increment from $q_t$ and the root embedding $E(r_t)$:
\begin{equation}
    \begin{aligned}
        x_t &= [\rho(q_t);\rho(E(r_t))],\\
        \widehat z_t
            &= q_t+B\,\operatorname{SiLU}(Ax_t+a)+b,\\
        z_t &=
        \begin{cases}
            q_t+\Delta_{j_t},
                & j_t\in\mathcal H_t,\ u_{j_t}=r_t,\\
            \widehat z_t, & \text{otherwise},
        \end{cases}
    \end{aligned}
    \label{eq:recycled_state}
\end{equation}
Here $\rho$ denotes root-mean-square normalization. For target hidden dimension $d$ and predictor bottleneck width $k$, $A\in\mathbb{R}^{k\times 2d}$ and $B\in\mathbb{R}^{d\times k}$ are the learned down- and up-projection matrices, with biases $a\in\mathbb{R}^{k}$ and $b\in\mathbb{R}^{d}$, respectively. Both branches estimate the post-root state from the current $q_t$ without an additional target-backbone forward pass.

The transition history is updated after each verification pass using target-computed hidden states, including those from rejected branches. Each state increment is measured relative to its parent state, using $q_t$ for the root.

\noindent\textbf{Residual predictor training.}
We train the 1.2M-parameter predictor on target-generated trajectories from 384 images per target, keeping the target VLM frozen. Its objective equally weights next-token cross-entropy and distribution matching to the target's true post-root state. Training supports and calibration details are provided in Appendix~\ref{sec:supp_head_calibration}.

\subsection{Visually Conditioned Candidate Reranking}

ReCue uses the estimated state $z_t$ to reassess a compact set of successor candidates. Let $W$ be the frozen target language head. We combine candidates from the already available root logits $Wq_t$ with those of the selected historical source, and score them using $z_t$:
\begin{equation}
    \begin{aligned}
        \mathcal V_t
            &= \operatorname{Top}_{128}(Wq_t)\cup\mathcal C_{j_t},\\
        \ell_t(v) &= (Wz_t)_v,\qquad v\in\mathcal V_t.
    \end{aligned}
    \label{eq:candidate_readout}
\end{equation}
The pool contains at most 256 distinct tokens; $\mathcal C_{j_t}=\emptyset$ if no source is available. Only the corresponding rows of $W$ are needed for post-root scoring. We rank candidates by $\ell_t$ and, for a local historical source, fuse this ranking with its cached successor ranks using equal-weight reciprocal-rank fusion. For bank sources, we use the state-based ranking alone.

The top-ranked candidates update the root's cached top-eight successor list, or \emph{Copy row}, yielding $\mathcal U_t$. Up to four proposals are inserted; when the row is target-generated, its original top-one prediction is preserved. This updated row guides drafting, while the original row is restored for verification.

\subsection{Contextual Draft-Tree Construction}

The updated root row $\mathcal U_t$ guides two complementary sources of draft paths. First, a request-local text index retrieves continuations from the prompt and accepted output, using ordinary suffix matching and one-token-shifted queries seeded by the proposed successors~\cite{liu2026logitspec}. Second, cached Copy rows provide additional successor predictions, using parent--current token pairs at deeper nodes when available. Retrieved paths are merged by shared prefixes, and remaining nodes are added by best-first Copy expansion using cached ranks and previously verified logits. We construct a tree of at most $N=64$ nodes, including the root, and verify it in a single target-model forward pass (Sec.~\ref{sec:prelim_tree}).

After verification, target predictions refresh the Copy rows, accepted tokens update the text index, and the last accepted node provides $q_{t+1}$ and $r_{t+1}$ for the next round.

\section{Experiments}
\label{sec:experiments}

\begin{table*}[tp]
    \centering
    \caption{Image-conditioned generation under natural EOS. Speedup is relative to paired AR decode throughput. Pooled combines the five datasets in this table. Bold marks the best reported value within each model.}
    \label{tab:recue_image}
    \small
    \setlength{\tabcolsep}{2.5pt}
    \begin{tabularx}{\textwidth}{l*{6}{>{\centering\arraybackslash}X}}
        \toprule
         & \textbf{MM-Vet} & \textbf{TextVQA} & \textbf{ScienceQA} & \textbf{CharXiv} & \textbf{MMMU-Pro} & \textbf{Pooled} \\
        \cmidrule(lr){2-2} \cmidrule(lr){3-3} \cmidrule(lr){4-4} \cmidrule(lr){5-5} \cmidrule(lr){6-6} \cmidrule(lr){7-7}
        \textbf{Method} & \recueMetricHeader & \recueMetricHeader & \recueMetricHeader & \recueMetricHeader & \recueMetricHeader & \recueMetricHeader \\
        \midrule[0.8pt]
        \multicolumn{7}{@{}l}{\textbf{Qwen2.5-VL-7B}} \\
        \recueMethodGroup{7}{Draft-model baselines}

        EAGLE-2 & \recueMetricPair{3.59}{1.88\texttimes} & \recueMetricPair{3.39}{1.67\texttimes} & \recueMetricPair{3.58}{1.91\texttimes} & \recueMetricPair{\textbf{3.82}}{1.93\texttimes} & \recueMetricPair{3.87}{2.05\texttimes} & \recueMetricPair{3.73}{1.96\texttimes} \\
        Medusa & \recueMetricPair{1.83}{1.36\texttimes} & \recueMetricPair{1.77}{1.26\texttimes} & \recueMetricPair{1.84}{1.33\texttimes} & \recueMetricPair{1.88}{1.36\texttimes} & \recueMetricPair{1.93}{1.43\texttimes} & \recueMetricPair{1.89}{1.39\texttimes} \\
        ViSpec & \recueMetricPair{3.33}{2.00\texttimes} & \recueMetricPair{3.21}{1.85\texttimes} & \recueMetricPair{3.48}{2.04\texttimes} & \recueMetricPair{3.44}{2.00\texttimes} & \recueMetricPair{3.49}{2.08\texttimes} & \recueMetricPair{3.44}{2.04\texttimes} \\
        HiViS & \recueMetricPair{\textbf{3.60}}{2.17\texttimes} & \recueMetricPair{\textbf{3.48}}{2.06\texttimes} & \recueMetricPair{\textbf{3.83}}{2.33\texttimes} & \recueMetricPair{3.74}{2.24\texttimes} & \recueMetricPair{3.81}{2.33\texttimes} & \recueMetricPair{\textbf{3.75}}{2.28\texttimes} \\
        EAGLE-3 & \recueMetricPair{2.70}{1.82\texttimes} & \recueMetricPair{3.17}{2.11\texttimes} & \recueMetricPair{2.85}{1.98\texttimes} & \recueMetricPair{2.87}{1.94\texttimes} & \recueMetricPair{2.22}{1.56\texttimes} & \recueMetricPair{2.48}{1.72\texttimes} \\
        \recueMethodRule
        \recueMethodGroup{7}{Model-free baselines}
        Token Recycling & \recueMetricPair{2.45}{2.20\texttimes} & \recueMetricPair{2.20}{1.94\texttimes} & \recueMetricPair{2.52}{2.24\texttimes} & \recueMetricPair{2.40}{2.14\texttimes} & \recueMetricPair{2.78}{2.52\texttimes} & \recueMetricPair{2.60}{2.34\texttimes} \\
        RACER & \recueMetricPair{2.37}{2.10\texttimes} & \recueMetricPair{1.98}{1.73\texttimes} & \recueMetricPair{2.53}{2.25\texttimes} & \recueMetricPair{2.65}{2.34\texttimes} & \recueMetricPair{2.92}{2.62\texttimes} & \recueMetricPair{2.65}{2.36\texttimes} \\
        RECAP & \recueMetricPair{2.75}{2.36\texttimes} & \recueMetricPair{2.36}{2.00\texttimes} & \recueMetricPair{3.01}{2.49\texttimes} & \recueMetricPair{3.14}{2.66\texttimes} & \recueMetricPair{3.55}{3.08\texttimes} & \recueMetricPair{3.16}{2.71\texttimes} \\
        \rowcolor{recueTableHighlight} \textbf{ReCue} {\footnotesize (1.2M head)} & \recueMetricPair{3.08}{\textbf{2.63\texttimes}} & \recueMetricPair{2.81}{\textbf{2.38\texttimes}} & \recueMetricPair{3.20}{\textbf{2.77\texttimes}} & \recueMetricPair{3.65}{\textbf{3.13\texttimes}} & \recueMetricPair{\textbf{3.94}}{\textbf{3.43\texttimes}} & \recueMetricPair{3.53}{\textbf{3.05\texttimes}} \\
        \midrule[0.8pt]
        \multicolumn{7}{@{}l}{\textbf{Qwen3-VL-8B}} \\
        \recueMethodGroup{7}{Draft-model baselines}
        DSpark & \recueMetricPair{2.80}{2.22\texttimes} & \recueMetricPair{2.79}{2.21\texttimes} & \recueMetricPair{2.62}{2.14\texttimes} & \recueMetricPair{3.87}{\textbf{3.09\texttimes}} & \recueMetricPair{2.84}{2.27\texttimes} & \recueMetricPair{3.03}{2.42\texttimes} \\
        JetSpec & \recueMetricPair{2.56}{2.11\texttimes} & \recueMetricPair{2.55}{2.10\texttimes} & \recueMetricPair{2.53}{2.01\texttimes} & \recueMetricPair{3.45}{2.85\texttimes} & \recueMetricPair{2.56}{2.12\texttimes} & \recueMetricPair{2.75}{2.26\texttimes} \\
        DFlash2 & \recueMetricPair{2.83}{2.12\texttimes} & \recueMetricPair{2.87}{2.17\texttimes} & \recueMetricPair{2.78}{2.16\texttimes} & \recueMetricPair{\textbf{3.88}}{2.95\texttimes} & \recueMetricPair{2.86}{2.20\texttimes} & \recueMetricPair{3.07}{2.34\texttimes} \\

        DFlash & \recueMetricPair{2.57}{2.13\texttimes} & \recueMetricPair{2.57}{2.13\texttimes} & \recueMetricPair{2.53}{2.12\texttimes} & \recueMetricPair{3.46}{2.87\texttimes} & \recueMetricPair{2.58}{2.14\texttimes} & \recueMetricPair{2.76}{2.29\texttimes} \\
        EAGLE-3 & \recueMetricPair{\textbf{3.28}}{2.25\texttimes} & \recueMetricPair{\textbf{3.00}}{2.04\texttimes} & \recueMetricPair{\textbf{3.49}}{2.43\texttimes} & \recueMetricPair{3.77}{2.58\texttimes} & \recueMetricPair{\textbf{3.36}}{2.30\texttimes} & \recueMetricPair{\textbf{3.43}}{2.35\texttimes} \\
        \recueMethodRule
        \recueMethodGroup{7}{Model-free baselines}
        Token Recycling & \recueMetricPair{2.38}{1.98\texttimes} & \recueMetricPair{2.29}{1.92\texttimes} & \recueMetricPair{2.41}{2.02\texttimes} & \recueMetricPair{2.74}{2.31\texttimes} & \recueMetricPair{2.55}{2.15\texttimes} & \recueMetricPair{2.54}{2.13\texttimes} \\
        RACER & \recueMetricPair{2.38}{1.96\texttimes} & \recueMetricPair{2.15}{1.79\texttimes} & \recueMetricPair{2.41}{1.97\texttimes} & \recueMetricPair{2.89}{2.41\texttimes} & \recueMetricPair{2.68}{2.24\texttimes} & \recueMetricPair{2.61}{2.17\texttimes} \\
        RECAP & \recueMetricPair{2.68}{2.15\texttimes} & \recueMetricPair{2.54}{2.06\texttimes} & \recueMetricPair{2.86}{2.31\texttimes} & \recueMetricPair{3.42}{2.78\texttimes} & \recueMetricPair{3.10}{2.52\texttimes} & \recueMetricPair{3.03}{2.46\texttimes} \\
        \rowcolor{recueTableHighlight} \textbf{ReCue} {\footnotesize (1.2M head)} & \recueMetricPair{2.99}{\textbf{2.41\texttimes}} & \recueMetricPair{2.81}{\textbf{2.27\texttimes}} & \recueMetricPair{3.03}{\textbf{2.49\texttimes}} & \recueMetricPair{3.76}{3.07\texttimes} & \recueMetricPair{3.30}{\textbf{2.70\texttimes}} & \recueMetricPair{3.29}{\textbf{2.68\texttimes}} \\
        \midrule[0.8pt]
        \multicolumn{7}{@{}l}{\textbf{LLaVA-NeXT Vicuna-7B}} \\
        \recueMethodGroup{7}{Draft-model baselines}
        DREAM-S & \recueMetricPair{2.69}{1.48\texttimes} & \recueMetricPair{2.06}{1.11\texttimes} & \recueMetricPair{2.80}{1.55\texttimes} & \recueMetricPair{2.44}{1.32\texttimes} & \recueMetricPair{2.51}{1.36\texttimes} & \recueMetricPair{2.53}{1.38\texttimes} \\
        DREAM & \recueMetricPair{2.79}{1.49\texttimes} & \recueMetricPair{2.37}{1.28\texttimes} & \recueMetricPair{2.99}{1.62\texttimes} & \recueMetricPair{2.73}{1.46\texttimes} & \recueMetricPair{2.62}{1.40\texttimes} & \recueMetricPair{2.71}{1.45\texttimes} \\
        EAGLE-2 & \recueMetricPair{3.12}{1.85\texttimes} & \recueMetricPair{3.45}{2.01\texttimes} & \recueMetricPair{2.88}{1.71\texttimes} & \recueMetricPair{2.65}{1.55\texttimes} & \recueMetricPair{2.29}{1.36\texttimes} & \recueMetricPair{2.69}{1.59\texttimes} \\
        HiViS & \recueMetricPair{\textbf{4.14}}{\textbf{2.61\texttimes}} & \recueMetricPair{\textbf{4.02}}{\textbf{2.48\texttimes}} & \recueMetricPair{\textbf{4.21}}{\textbf{2.65\texttimes}} & \recueMetricPair{\textbf{4.23}}{2.64\texttimes} & \recueMetricPair{4.37}{2.74\texttimes} & \recueMetricPair{\textbf{4.23}}{2.65\texttimes} \\
        ViSpec & \recueMetricPair{4.03}{2.51\texttimes} & \recueMetricPair{4.00}{2.42\texttimes} & \recueMetricPair{3.98}{2.44\texttimes} & \recueMetricPair{4.13}{2.53\texttimes} & \recueMetricPair{4.15}{2.55\texttimes} & \recueMetricPair{4.08}{2.51\texttimes} \\
        \recueMethodRule
        \recueMethodGroup{7}{Model-free baselines}
        Token Recycling & \recueMetricPair{2.46}{1.88\texttimes} & \recueMetricPair{2.44}{1.86\texttimes} & \recueMetricPair{2.54}{1.95\texttimes} & \recueMetricPair{2.63}{2.01\texttimes} & \recueMetricPair{2.77}{2.14\texttimes} & \recueMetricPair{2.61}{2.00\texttimes} \\
        RACER & \recueMetricPair{2.44}{1.86\texttimes} & \recueMetricPair{2.39}{1.82\texttimes} & \recueMetricPair{2.71}{2.07\texttimes} & \recueMetricPair{2.74}{2.10\texttimes} & \recueMetricPair{3.22}{2.48\texttimes} & \recueMetricPair{2.77}{2.12\texttimes} \\
        RECAP & \recueMetricPair{2.85}{2.12\texttimes} & \recueMetricPair{2.28}{1.70\texttimes} & \recueMetricPair{3.12}{2.32\texttimes} & \recueMetricPair{3.33}{2.49\texttimes} & \recueMetricPair{3.76}{2.83\texttimes} & \recueMetricPair{3.19}{2.39\texttimes} \\
        \rowcolor{recueTableHighlight} \textbf{ReCue} {\footnotesize (1.2M head)} & \recueMetricPair{3.18}{2.36\texttimes} & \recueMetricPair{3.30}{2.45\texttimes} & \recueMetricPair{3.36}{2.52\texttimes} & \recueMetricPair{3.84}{\textbf{2.87\texttimes}} & \recueMetricPair{\textbf{4.80}}{\textbf{3.62\texttimes}} & \recueMetricPair{3.81}{\textbf{2.84\texttimes}} \\
        \bottomrule
    \end{tabularx}
\end{table*}

\subsection{Experimental Setup}

We evaluate Qwen2.5-VL-7B-Instruct~\cite{bai2025qwen25vl}, Qwen3-VL-8B-Instruct~\cite{bai2025qwen3vl}, and LLaVA-NeXT Vicuna-7B~\cite{liu2024llavanext}. Image evaluation uses 100 requests per benchmark from MM-Vet~\cite{yu2024mmvet}, TextVQA~\cite{singh2019textvqa}, ScienceQA~\cite{lu2022scienceqa}, CharXiv~\cite{wang2024charxiv}, and MMMU-Pro~\cite{yue2025mmmupro}. Video evaluation uses two disjoint sets of 114 and 157 requests drawn from VideoDetailCaption~\cite{lmmslab2024vdc}, MVBench~\cite{li2024mvbench}, Video-MME~\cite{fu2025videomme}, MLVU~\cite{zhou2025mlvu}, and LongVideoBench~\cite{wu2024longvideobench}. Qwen models process up to 32 video frames, while LLaVA-NeXT uses four-frame storyboards.

All methods use paired inputs and target checkpoints, BF16 precision, batch size one, and greedy decoding on NVIDIA A40 GPUs. Generation stops at EOS or a maximum of 1,024 new tokens, subject to KV-cache capacity. We report mean accepted tokens (MAT) as
$\sum_i g_i / \sum_i r_i$, where $g_i$ counts emitted tokens
including roots and $r_i$ counts target verification rounds
for request $i$. MAT measures how many tokens generation
advances per verification round. Speedup uses pooled decode
throughput relative to paired autoregressive (AR) decoding.

\subsection{Baselines}

We compare ReCue against model-free prediction-recycling methods and learned drafters. Model-free baselines include Token Recycling~\cite{luo2025tokenrecycling}, RACER~\cite{zhang2026racer}, and RECAP~\cite{zhu2026recap}, which progressively extend cached prediction reuse with continuation retrieval and persistent memories. All use a maximum of 64 draft-tree nodes.

Sequential-drafter baselines include EAGLE-2/3~\cite{li2024eagle2,li2025eagle3}, ViSpec~\cite{kang2025vispec}, HiViS~\cite{xie2026hivis}, Medusa~\cite{cai2024medusa}, DREAM/DREAM-S~\cite{hu2025dream,liu2026dreams}, and Qwen3-VL-compatible DFlash, DFlash2, DSpark, and JetSpec variants~\cite{chen2026dflash,inco2026dflash2,cheng2026dspark,hu2026jetspec,veladraft2026}. We additionally evaluate video-specific methods SpecVLM~\cite{ji2025specvlm} and ParallelVLM~\cite{kong2026parallelvlm}. We use available compatible checkpoints or implementations; model-specific configurations are detailed in the supplementary material. ReCue keeps the target frozen and trains only a 1.2M-parameter residual head.

\subsection{Main Results}

\paragraph{Image decoding.}
Table~\ref{tab:recue_image} compares ReCue with representative prediction-recycling methods and learned drafters across three VLMs and five image benchmarks. ReCue achieves the highest pooled speedup on all three targets, reaching 3.05$\times$, 2.68$\times$, and 2.84$\times$ relative to autoregressive decoding. These results demonstrate consistent decoding gains across different target architectures and visual understanding tasks.

\noindent\textbf{Prediction-recycling baselines.}
RACER combines cached predictions with exact text-continuation retrieval, while RECAP further exploits request-local and persistent prediction memories. Although their cached predictions originate from target-model computations, the retrieved successor rankings are not explicitly adapted to the current visual context. ReCue instead estimates a visually conditioned post-root state and uses the frozen target language head to rerank reusable candidates. These candidates subsequently guide contextual retrieval and Copy-based tree expansion. Compared with RECAP, ReCue improves pooled MAT by 8.6--19.4\% and speedup by 8.9--18.8\% across the three targets. The simultaneous gains in acceptance length and throughput suggest that recontextualizing historical predictions, together with improved tree construction, enables more effective use of the fixed verification budget.

\noindent\textbf{Sequence-conditioned drafters and drafting efficiency.}
Sequence-conditioned approaches such as EAGLE-3 and HiViS exploit target representations to predict future tokens, with HiViS additionally conveying visual information through target hidden states. In contrast, ReCue directly reuses target predictions and employs only a lightweight 1.2M-parameter residual head, evaluated once per decoding round, to estimate the post-root state. The frozen language head then scores a compact candidate pool, avoiding a separate autoregressive drafting model. This design offers a favorable trade-off between draft quality and computational overhead. For example, on Qwen3-VL-8B, EAGLE-3 achieves higher pooled MAT than ReCue (3.43 vs.\ 3.29), yet ReCue attains greater speedup (2.68$\times$ vs.\ 2.35$\times$). A similar pattern appears against HiViS on LLaVA-NeXT (4.23 vs.\ 3.81 MAT; 2.65$\times$ vs.\ 2.84$\times$ speedup), and is also observed in our video and SGLang evaluations. These results highlight the practical advantage of lightweight drafting: longer accepted sequences do not necessarily translate into faster decoding when draft-generation costs are considered.

\noindent\textbf{Task-dependent differences.}
The improvements are not uniform across all benchmarks. On Qwen3-VL CharXiv, DSpark slightly outperforms ReCue (3.09$\times$ vs.\ 3.07$\times$), while HiViS achieves higher throughput on LLaVA-NeXT MM-Vet, TextVQA, and ScienceQA. These cases indicate that learned drafters can retain advantages on particular model--task combinations. Conversely, ReCue performs particularly well on LLaVA-NeXT CharXiv and MMMU-Pro, achieving 2.87$\times$ and 3.62$\times$ speedup, respectively, compared with HiViS's 2.64$\times$ and 2.74$\times$. Overall, ReCue does not rely on universally longer accepted drafts, but achieves stronger pooled efficiency by combining context-adaptive prediction reuse with low-cost draft construction. We further investigate the role of visual conditioning through controlled experiments in Sec.~\ref{sec:visual_analysis}.

\begin{table}[t]
\centering
\caption{Video-conditioned decoding pooled across benchmarks.
Speedup is relative to paired AR.
Bold marks the best value per metric and target.}
\label{tab:video_generalization}
\small
\setlength{\tabcolsep}{1pt}
\renewcommand{\arraystretch}{1.03}
\begin{minipage}[t]{0.485\columnwidth}
\centering
\begin{tabular*}{\linewidth}{@{\extracolsep{\fill}}lrr@{}}
\toprule
\multicolumn{3}{c}{\textbf{Qwen2.5-VL-7B}} \\
\midrule
Method & MAT & Speedup \\
\midrule
SpecVLM      & \textbf{4.81} & 1.12$\times$ \\
ViSpec       & 2.91          & 1.74$\times$ \\
HiViS        & 3.68          & 2.22$\times$ \\
ParallelVLM  & 1.65          & 1.26$\times$ \\
RECAP        & 2.83          & 2.18$\times$ \\
\midrule
\textbf{ReCue} & 3.05 & \textbf{2.37$\times$} \\
\bottomrule
\end{tabular*}
\end{minipage}\hfill
\begin{minipage}[t]{0.485\columnwidth}
\centering
\begin{tabular*}{\linewidth}{@{\extracolsep{\fill}}lrr@{}}
\toprule
\multicolumn{3}{c}{\textbf{Qwen3-VL-8B}} \\
\midrule
Method & MAT & Speedup \\
\midrule
ParallelVLM     & 1.99          & 1.14$\times$ \\
EAGLE-3         & \textbf{3.32} & 2.40$\times$ \\
Token Recycling & 2.48          & 2.15$\times$ \\
RACER           & 2.42          & 2.09$\times$ \\
RECAP           & 2.89          & 2.46$\times$ \\
\midrule
\textbf{ReCue} & 3.02 & \textbf{2.57$\times$} \\
\bottomrule
\end{tabular*}
\end{minipage}
\end{table}

\paragraph{Generalization to video.}
ReCue's residual predictor is calibrated exclusively on image trajectories, without video-specific alignment. The same state-recycling and candidate-reranking mechanism applies to video-conditioned generation, where the target's hidden states encode information from multiple frames. As shown in Table~\ref{tab:video_generalization}, ReCue achieves pooled speedups of 2.37$\times$ and 2.57$\times$ over AR on Qwen2.5-VL-7B and Qwen3-VL-8B, respectively.

On Qwen2.5-VL, ReCue outperforms SpecVLM, ViSpec, HiViS, ParallelVLM, and RECAP in decode throughput. On Qwen3-VL, it also exceeds ParallelVLM, EAGLE-3, Token Recycling, and RACER, while improving over the strongest prediction-recycling baseline, RECAP (2.57$\times$ vs.\ 2.46$\times$). These results suggest that visual-context adaptation through the target's existing representations remains effective beyond static-image generation, without requiring an additional video-specific drafter.

\begin{table}[t]
\centering
\caption{
Single-request decoding in the optimized SGLang framework on Qwen3-VL-8B using A40 GPUs. Speedup is relative to paired autoregressive decoding under the same backend. Best results in each column are in bold.
}
\label{tab:sglang_main}
\small
\setlength{\tabcolsep}{5pt}
\begin{tabular}{lcccc}
\toprule
& \multicolumn{2}{c}{Image}
& \multicolumn{2}{c}{Video} \\
\cmidrule(lr){2-3}
\cmidrule(lr){4-5}
Method & MAT & Speedup & MAT & Speedup \\
\midrule
N-gram
    & 1.91 & 1.57$\times$
    & 2.23 & 1.55$\times$ \\
RACER
    & 2.56 & 1.93$\times$
    & 2.65 & 1.43$\times$ \\
EAGLE-3
    & \textbf{3.33} & 1.98$\times$
    & \textbf{3.50} & 1.81$\times$ \\
\textbf{ReCue}
    & 3.20 & \textbf{2.47$\times$}
    & 3.27 & \textbf{2.14$\times$} \\
\bottomrule
\end{tabular}
\end{table}

\paragraph{Optimized inference with SGLang.}
We integrate ReCue into SGLang and evaluate it against baselines with available SGLang implementations for Qwen3-VL-8B: N-gram, RACER, and EAGLE-3. Table~\ref{tab:sglang_main} reports image and video generation results under the same backend and verification budget. ReCue achieves 2.47$\times$ and 2.14$\times$ speedup over autoregressive decoding, respectively. Compared with the strongest reported baseline in each modality, this corresponds to approximately 25\% higher image throughput and 18\% higher video throughput. These results show that ReCue combines competitive MAT with low drafting overhead to achieve meaningful throughput gains even in an optimized inference backend.

\begin{table}[t]
\centering
\caption{
Module ablation on Qwen3-VL-8B using A40 GPUs. Entries are pooled MAT under natural EOS with a 64-node verification budget.
}
\label{tab:module_ablation}
\small
\setlength{\tabcolsep}{8pt}
\renewcommand{\arraystretch}{1}
\begin{tabular}{lcc}
\toprule
Configuration & Image & Video \\
\midrule
RACER & 2.61 & 2.42 \\
\quad + State recycling & 2.85 & 2.66 \\
\quad + Tree expansion & 3.07 & 2.75 \\
\midrule
ReCue w/ native Copy & 2.97 & 2.75 \\
ReCue w/ native retrieval & 3.21 & 2.99 \\
\textbf{ReCue} & \textbf{3.29} & \textbf{3.02} \\
\bottomrule
\end{tabular}
\end{table}

\paragraph{Module ablation.}
Table~\ref{tab:module_ablation} examines the contributions of ReCue's main components on Qwen3-VL-8B.
\noindent\textbf{State recycling and candidate reranking.}
Adding state recycling to RACER improves pooled MAT by 9.2\% on images and 9.9\% on videos. This variant estimates the post-root state and uses the frozen language head to rerank cached successor candidates, while retaining RACER's original tree construction. The improvements indicate that adapting historical predictions to the current target state is beneficial even without the proposed contextual tree expansion.
\noindent\textbf{Contextual draft-tree construction.}
Independently introducing the proposed tree expansion to RACER improves MAT by 17.4\% and 13.6\% on images and videos, respectively. This demonstrates that contextual Copy expansion and continuation retrieval improve the utilization of the fixed verification budget, even without state-based reranking. Combining both components in the full ReCue framework yields the highest MAT, improving over RACER by 26.0\% and 24.8\% across the two modalities.

We further examine the two mechanisms within tree construction. Replacing contextual Copy with its native policy reduces MAT by 9.7\% on images and 8.9\% on videos, whereas reverting to native retrieval causes smaller reductions of 2.5\% and 1.1\%. These results suggest that Copy-based expansion contributes more substantially to the overall gains, while the enhanced retrieval strategy provides complementary improvements. In full ReCue, we use the estimated post-root state to rerank the root’s successor candidates, and the updated successor list guides both Copy expansion and continuation retrieval.

\begin{table}[t]
\centering
\caption{Isolating historical transfer and residual prediction under paired image-prefix replays. Entries are paired increases in emitted tokens per verification with 95\% request-bootstrap intervals.}
\label{tab:state_components_main}
\small
\setlength{\tabcolsep}{3pt}
\renewcommand{\arraystretch}{1}
\begin{tabularx}{\columnwidth}{@{}l*{3}{>{\centering\arraybackslash}X}@{}}
\toprule
Component & Qwen2.5 & Qwen3 & LLaVA \\
\midrule
Historical transfer & \shortstack{+0.071\\{\scriptsize [0.053, 0.091]}} & \shortstack{+0.061\\{\scriptsize [0.051, 0.070]}} & \shortstack{+0.065\\{\scriptsize [0.050, 0.080]}} \\
Predictor fallback & \shortstack{+0.122\\{\scriptsize [0.099, 0.148]}} & \shortstack{+0.109\\{\scriptsize [0.096, 0.124]}} & \shortstack{+0.111\\{\scriptsize [0.091, 0.133]}} \\
\bottomrule
\end{tabularx}
\end{table}

\paragraph{Ablating hidden-state estimation.}
We isolate the contributions of historical transfer and residual prediction by comparing three configurations: (1) ranking candidates directly with the current state $q_t$, (2) additionally reusing historical transitions on matched rounds, and (3) further enabling predictor fallback on unmatched rounds. We fix the prefix, root, candidate pool, and tree policy across configurations on 100 image requests per target.
As shown in Table~\ref{tab:state_components_main}, historical transfer improves emitted tokens per verification by 0.061--0.071 over current-state ranking, while predictor fallback provides an additional gain of 0.109--0.122. All six 95\% request-bootstrap intervals exclude zero. Moreover, using the predictor alone improves throughput by 4.3--7.7\% over current-state ranking in complete image decoding.

\subsection{Case Study}
Figure~\ref{fig:recue_cost_benefit} relates drafting cost to net decoding gains on Qwen2.5-VL-7B using A40 GPUs. In the MMMU-Pro case, state matching, prediction, candidate scoring, and tree construction together occupy 4.25\% of decode time. Across the full image suite, combining state recycling with contextual tree expansion increases MAT from 2.65 to 3.53 and throughput from $2.36\times$ to $3.05\times$ AR, yielding a 29.2\% net improvement.

\subsection{Visual Conditioning Analysis}
\label{sec:visual_analysis}

\begin{table}[t]
\centering\small
\setlength{\tabcolsep}{4pt}
\renewcommand{\arraystretch}{1.06}
\caption{Visual evidence from retrieval to acceptance. Arrows show Frozen $\to$ ReCue for OCR and Historical $\to$ ReCue for real-task ranking. $\Delta A$ is the mean acceptance gain per round; brackets give paired 98.75\% cluster intervals.}
\label{tab:visual_chain}
\begin{tabularx}{\columnwidth}{@{}X>{\centering\arraybackslash}p{.27\columnwidth}>{\centering\arraybackslash}p{.27\columnwidth}@{}}
\toprule
Measurement & Qwen2.5-VL & Qwen3-VL\\
\midrule
\multicolumn{3}{@{}l}{\textit{Controlled OCR (400 cases / target)}}\\
Top-4 coverage (\%) & 40 $\to$ 100 & 40 $\to$ 80 \\
\midrule
\multicolumn{3}{@{}l}{\textit{Real-task replay (100 images / target)}}\\
Pool recall (\%) & 91.58 & 91.50 \\
Top-4 in pool (\%) & 65.56 $\to$ 77.34 & 64.47 $\to$ 75.26 \\
$\Delta A$ over history & \shortstack[c]{+0.121\\[-1pt][0.089, 0.155]} & \shortstack[c]{+0.121\\[-1pt][0.102, 0.141]} \\
$\Delta A$ vs. swapped image & \shortstack[c]{+0.028\\[-1pt][0.012, 0.045]} & \shortstack[c]{+0.028\\[-1pt][0.016, 0.040]} \\
\midrule
\multicolumn{3}{@{}l}{\textit{Online decoding (100 images / target)}}\\
MAT gain vs. RACER & \textbf{+34.2\%} & \textbf{+28.8\%} \\
MAT gain vs. RECAP & \textbf{+11.0\%} & \textbf{+7.5\%} \\
\bottomrule
\end{tabularx}
\end{table}

We test visual grounding with a controlled OCR task (Fig.~\ref{fig:recue_visual_state}a). We change one digit in a chart legend while keeping the prompt and prefix fixed. Across 400 cases per target, ReCue includes the edited digit among four proposals in 100\%/80\% of cases (Qwen2.5-VL/Qwen3-VL), compared with 40\% for the frozen ranking (Fig.~\ref{fig:recue_visual_state}b).

On 100 natural images per target, reranking the same candidates increases acceptance by 0.121 tokens per round over historical ranking on both models. With online cache updates, ReCue improves MAT over RACER by 34.2\%/28.8\% and RECAP by 11.0\%/7.5\%. Table~\ref{tab:visual_chain} connects coverage to accepted tokens; detailed controls are provided in the supplement.

\section{Conclusion}

ReCue brings the target model's visual awareness into cache-based draft-tree construction.
By recycling visually conditioned hidden states, it guides successor selection, continuation retrieval, and tree expansion with the current multimodal context.
Across three target models, ReCue achieves pooled decode speedups of $2.68$--$3.05\times$ on images and $2.37$--$2.57\times$ on video over autoregressive decoding, with gains also demonstrated in SGLang.
Controlled visual interventions and image-swap replays show how current visual evidence shapes successor priorities and accepted paths.

\begin{figure}[t]
    \centering
    \includegraphics[width=\columnwidth]{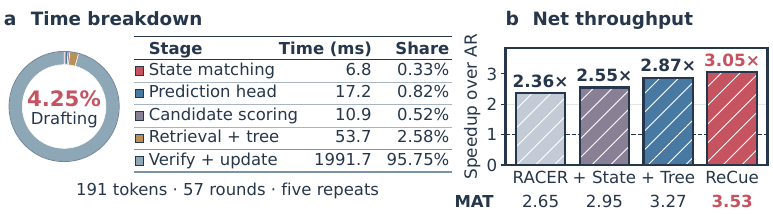}
    \caption{ReCue on Qwen2.5-VL-7B: (a) decode-time breakdown; (b) module ablation.}
    \label{fig:recue_cost_benefit}
\end{figure}

\FloatBarrier
\section*{LLM Usage Statement}

We used GPT models for language editing, cross-checking experimental records, and supplementary reproducibility checks, including screening the predictor's training and calibration images for overlap with evaluation inputs, repeated timing and statistical analysis, and sanity checks of tree verification, cache behavior, and agreement with greedy autoregressive decoding. The authors take full responsibility for the manuscript and all reported results.

{\small
\bibliographystyle{plainnat}
\bibliography{references}
}

\clearpage
\appendix
\twocolumn[{\centering\Large\bfseries Appendix\par\vspace{12pt}}]

\section{Experimental Details}
\label{sec:supp_experimental_details}

\subsection{Models and Evaluation Benchmarks}

\paragraph{Target models.}
We evaluate Qwen2.5-VL-7B-Instruct, Qwen3-VL-8B-Instruct, and LLaVA-NeXT Vicuna-7B with frozen target weights. ReCue trains an approximately 1.2M-parameter residual predictor on target-generated image trajectories (Sec.~\ref{sec:supp_head_calibration}).

\paragraph{Image benchmarks.}
Each target uses 100 requests from each of five image benchmarks: MM-Vet, TextVQA, ScienceQA, CharXiv, and MMMU-Pro. MM-Vet retains its original questions. ScienceQA uses image-based natural-science questions and asks for an option letter with a brief explanation.

\paragraph{Video benchmarks.}
Video evaluation uses two disjoint sets of 114 and 157 requests from VideoDetailCaption, MVBench, Video-MME, MLVU, and LongVideoBench. VideoDetailCaption retains its prompts; the other benchmarks share a detailed-description instruction. Qwen2.5-VL and Qwen3-VL use native video inputs with up to 32 frames; LLaVA-NeXT receives four-frame, temporally ordered storyboards.

\paragraph{Additional evaluation sets.}
Our SGLang image evaluation uses GQA~\cite{hudson2019gqa}, TextVQA, COCO Captions~\cite{chen2015cococaptions}, CharXiv, and MMMU-Pro.

\subsection{Decoding Settings and Metrics}

\paragraph{Generation protocol.}
Single-request comparisons match inputs and target checkpoints, with BF16 target weights, batch size one, and greedy decoding. Generation stops at natural EOS or 1,024 new tokens unless a lower cap is specified in Table~\ref{tab:supp_protocols} or required by the KV capacity. Sections~\ref{sec:supp_visual_evidence} and~\ref{sec:supp_system} describe the visual and system measurements.

\paragraph{Evaluation metrics.}
We report mean accepted tokens (MAT) and decoding speedup. MAT measures the average number of emitted tokens per target verification round, including the target-selected root:
\begin{equation}
    \mathrm{MAT}
    =
    \frac{\sum_i g_i}{\sum_i r_i},
\end{equation}
where $g_i$ and $r_i$ denote the number of emitted tokens and target verification rounds for request $i$, respectively.

Decoding speedup is the ratio of pooled decoded-token throughput to that of paired autoregressive decoding (AR). Throughput is total generated tokens divided by total decode time. We measure candidate recall and acceptance in fixed-prefix and visual-intervention experiments, and throughput during complete online generation.

\begin{table}[!htbp]
\centering\small
\setlength{\tabcolsep}{2.5pt}
\renewcommand{\arraystretch}{1.03}
\caption{Evaluation workloads and generation caps.}
\label{tab:supp_protocols}
\begin{tabular*}{\columnwidth}{@{\extracolsep{\fill}}p{101pt}p{85pt}r@{}}
\toprule
Experiment & Requests per target & Cap \\
\midrule
\multicolumn{3}{@{}l}{\textit{A40}}\\
Native main tables & 500 images; 271 videos & 1,024 \\
Fixed-prefix state probes & 100 images; 50 videos & 1,024 \\
Root updates and readouts & 88 images & 1,024 \\
Online root updates & 88 images $\times$ 3 repeats & 1,024 \\
Video estimation policies & 100 videos $\times$ 3 repeats & 1,024 \\
Official conversations & 100 sessions; 223 turns & 1,024 \\
AVSD dialogues & 32 videos; 320 turns & 1,024 \\
Multi-turn reranking & 24 videos $\times$ 3 turns & 1,024 \\
Single-request SGLang & 500 images; 271 videos & 1,024 \\
COCO query intervention & 100 images & 1,024 \\
Repeated native timing & 400 images $\times$ 3 repeats & 256 \\
RECAP controls & 100 images $\times$ 3 repeats & 256 \\
\bottomrule
\end{tabular*}
\end{table}

\subsection{Baseline Implementations}

\paragraph{Model-free baselines.}
We evaluate Token Recycling, RACER, and RECAP with at most 64 draft-tree nodes. RACER uses its native drafting policy~\cite{zhang2026racer} on the shared target backend.

\paragraph{Qwen2.5-VL and LLaVA-NeXT baselines.}
For Qwen2.5-VL-7B and LLaVA-NeXT Vicuna-7B, we evaluate ViSpec, HiViS, and visual EAGLE-2 checkpoints~\cite{shen2026mmspec,xie2026hivis}.

Qwen2.5-VL comparisons also include Medusa with MMSpec weights~\cite{shen2026mmspec} and EAGLE-3 with the Rayzl visual checkpoint~\cite{rayzl2025qwen25eagle3}. LLaVA-NeXT comparisons include DREAM and DREAM-S. HiViS, DREAM, and DREAM-S~\cite{xie2026hivis,hu2025dream,liu2026dreams} train on subsets of LLaVA-Mix665K, which includes GQA, TextVQA, and COCO.

\paragraph{Qwen3-VL baselines.}
For Qwen3-VL-8B, we use the MNN EAGLE-3 checkpoint. We additionally evaluate DFlash, DSpark, DFlash2, and JetSpec using the independently trained multimodal checkpoints released through \href{https://github.com/LUOseu/VelaDraft}{VelaDraft}~\cite{veladraft2026}.

VelaDraft reconstructs the DFlash2 training objective. With eight-position drafting blocks, DSpark verifies nine positions including the anchor; DFlash, DFlash2, and JetSpec verify eight. JetSpec uses causal attention within the block.

\paragraph{Video-specific baselines.}
SpecVLM uses a Qwen2.5-VL-3B-Instruct drafter with the Qwen2.5-VL-7B target and a 64-node tree. Our ParallelVLM reimplementation~\cite{kong2026parallelvlm} places target and drafter on separate NVIDIA A40 GPUs. Qwen2.5-VL-7B uses a second 7B instance as the drafter; Qwen3-VL-8B uses Qwen3-VL-2B-Instruct. Both use a pruning ratio of 0.9 and early layer 2, with draft windows of 2 and 4, respectively.

\subsection{Timing Protocol}

Native decode timing excludes visual encoding and prompt prefill. Qwen2.5-VL EAGLE-2 starts after target prefill and includes draft initialization.

Single-request SGLang disables CUDA Graphs and scheduling overlap and uses a request-local SSSD cache. We time from the first streamed output to completion, excluding the first output's tokens from the throughput numerator, with the same procedure for AR. Qwen video uses 16 frames; LLaVA-NeXT uses four-frame storyboards. The latency evaluation in Table~\ref{tab:supp_sglang_latency} uses one pass per target--modality pair. Section~\ref{sec:supp_concurrent_serving} specifies concurrent serving.

ParallelVLM reuses paired AR measurements for Qwen3-VL; Qwen2.5-VL times AR in the target process.

\subsection{Hardware and Software}
\label{sec:supp_hardware_software}

We use a server with eight NVIDIA A40 GPUs (48\,GB each), dual Intel Xeon Platinum 8383C CPUs (80 physical cores in total), and Ubuntu 20.04. Single-request ReCue decoding uses one GPU. Our implementations use Python 3.10. ReCue's native backend uses PyTorch 2.5.1 (CUDA 12.1), Transformers 4.57.0, and Triton 3.1.0. The SGLang backend uses PyTorch 2.13.0 (CUDA 12.6), Transformers 5.12.1, and FlashInfer 0.6.17.

\section{Residual-Head Calibration}
\label{sec:supp_head_calibration}

\paragraph{Data and trajectories.}
All targets share 384 training images (96 each from TextVQA, DocVQA, ChartQA, and AI2D) and 32 development images. Target-generated trajectories, capped at 128 tokens, provide pre-root and post-root final-layer states and target successors. Qwen2.5-VL, Qwen3-VL, and LLaVA-NeXT yield 25,169, 42,579, and 28,585 training transitions, respectively.

\paragraph{Predictor dimensions.}
For hidden width $d$ and bottleneck width $k$, the predictor uses $A\in\mathbb R^{k\times2d}$, $B\in\mathbb R^{d\times k}$, $a\in\mathbb R^k$, and $b\in\mathbb R^d$, totaling $3dk+k+d$ parameters (Tab.~\ref{tab:supp_head_configuration}). Retrieval uses top-128 successor lists and an at-most-256-token scoring pool.

\begin{table}[!htbp]
\centering
\small
\setlength{\tabcolsep}{3pt}
\renewcommand{\arraystretch}{1.07}
\caption{Residual-predictor dimensions and parameter counts.}
\label{tab:supp_head_configuration}
\begin{tabular*}{\columnwidth}{@{\extracolsep{\fill}}lrrr@{}}
\toprule
Target & $d$ & $k$ & Parameters \\
\midrule
Qwen2.5-VL-7B & 3,584 & 107 & 1,154,155 \\
Qwen3-VL-8B & 4,096 & 94 & 1,159,262 \\
LLaVA-NeXT-7B & 4,096 & 94 & 1,159,262 \\
\bottomrule
\end{tabular*}
\end{table}

\paragraph{Alignment objective.}
The residual predictor aligns the current visually conditioned target state with the state after the known root token. Cached target transitions provide current state $q$, root $r$, true post-root state $h^+$, and successor $y$. From normalized inputs, the predictor estimates $\widehat z=q+f_\theta(q,E(r))$. The same frozen language head $W$ reads $\widehat z$ and $h^+$ on support $\Omega$:
\begin{equation}
    p_\theta^\Omega=\operatorname{softmax}_\Omega(W\widehat z),
    \qquad
    p_*^\Omega=\operatorname{softmax}_\Omega(Wh^+).
\end{equation}
We minimize
\begin{equation}
    \mathcal L=\tfrac12\operatorname{CE}(y,p_\theta^\Omega)
        +\tfrac12\operatorname{CE}(p_*^\Omega,p_\theta^\Omega).
\end{equation}
The first term promotes the target successor; the second aligns predicted and target continuation distributions through the frozen language head. For Qwen2.5-VL, $\Omega$ is the deduplicated recorded candidate pool, padded to 256 entries with padding masked from the softmax. We optimize on the 25,169 transitions whose successor labels occur in these pools. Qwen3-VL and LLaVA-NeXT use the full vocabulary for $\Omega$; all three targets use compact candidate readout at inference.

\paragraph{Optimization.}
We train the residual head on cached states with the target and language head frozen. AdamW uses learning rate $10^{-3}$, weight decay 0.01, cosine decay, and gradient clipping at 5. Batch size is 64 for Qwen2.5-VL and 32 otherwise. Image-balanced transition weights give each training image equal weight; checkpoint selection minimizes image-averaged development cross-entropy. Runs finish after 37, 8, and 11 epochs, selecting epochs 27, 3, and 6 for Qwen2.5-VL, Qwen3-VL, and LLaVA-NeXT, respectively.

Cached-state optimization and per-epoch development evaluation take 47.7, 65.3, and 31.3 seconds on one GPU in target order. These fitting times exclude trajectory collection, bank construction, and initial teacher-distribution preparation.

\paragraph{Calibration and evaluation separation.}
Source banks, Copy priors, and rank profiles use target-generated calibration trajectories. Rank profiles use 384 calibration images for Qwen2.5-VL and LLaVA-NeXT, and 64 of the 384 training images for Qwen3-VL. Qwen2.5's bank also uses eight calibration images; its exclusion list therefore includes 384 training, 32 development, and eight bank-calibration images. Screening against main-table and SGLang inputs finds no matches by file hash, decoded-RGB hash, normalized $256\times256$ RGB hash, or 64-bit difference-hash distance at most two. The COCO captioning cohort uses the same exclusion list.

\section{Implementation Details}
\label{sec:supp_prediction_recycling}

ReCue uses the target's existing multimodal computation to guide a cache-built draft tree. The current state $q_t$ encodes the visual input and verified text and predicts root $r_t$; selecting its successors requires an estimate of the state after $r_t$. A compatible historical transition or the aligned residual predictor supplies this proposal state. The frozen target language head then reranks current and retrieved candidates. Selected successors guide root edges and queries for longer continuations, while cached target predictions fill the remaining tree budget. Verification refreshes the states and predictions available for the next round.

\subsection{Post-Root State Estimation}

Retrieved candidates come from request-local history $\mathcal H_t$ and a frozen bank $\mathcal B(r_t)$. Up to 256 local records store tokens $u_j$, normalized pre-token keys from $h_j^-$, raw increments $\Delta_j=h_j^+-h_j^-$, and ordered top-128 successors $\mathcal C_j$ from $Wh_j^+$. Here $h_j^+$ is the post-token state and $W$ the frozen target language head. Each bank holds 980 calibration records with matching keys and 128 successor IDs, indexed by root with at most 16 records per token. Qwen3 samples these records across training images.

At request start, the history is cleared and seeded from eligible text after the prompt's final visual marker, keeping the most recent positions that fit. A circular cursor overwrites the oldest inserted records. The history, contextual Copy rows, and accepted-text index reset per request; the bank and offline Copy priors persist.

With token embedding $E$, we score each available source as
\begin{equation}
    \begin{aligned}
        s_t(j)
            &= \cos(q_t,h_j^-)
               +\cos(E(r_t),E(u_j))-\kappa_j,\\
        j_t
            &= \operatorname*{arg\,max}^{\mathrm{lex}}_{
                j\in\mathcal H_t\cup\mathcal B(r_t)}
               \bigl(\mathbf{1}[u_j=r_t],\,s_t(j)\bigr).
    \end{aligned}
    \label{eq:state_source}
\end{equation}
Exact-root matches take precedence. The penalty $\kappa_j=10^{-7}a_j$ uses insertion age for local records and $256+s$ for bank slot $s\in\{0,\ldots,15\}$, favoring recent local entries and fixing bank tie order. We set $j_t=\bot$ when no source exists.

A selected same-root local source gives $z_t=q_t+\Delta_{j_t}$; remaining rounds use $z_t=q_t+f_\theta(q_t,E(r_t))$, with the residual map aligned as in Sec.~\ref{sec:supp_head_calibration}. Both estimates retain the current $q_t$ and share candidate retrieval and readout. EAGLE~\cite{li2024eagle} uses an advanced token sequence for future-feature prediction, and HiViS~\cite{xie2026hivis} supplies fused target representations to an autoregressive drafter. ReCue uses its post-root estimate for one root-level candidate readout, then expands the tree through cached predictions and text. Target KV and verification use target-computed states.

\subsection{Visually Conditioned Candidate Reranking}

The frozen language head rescores the union of the current top-128 candidates and the selected source's successor list in the estimated post-root state. Let $\pi_z(v)$ be the rank of candidate $v$ according to $\ell_t(v)$, and $\pi_C(v)$ its rank in the selected historical successor list, with $\pi_C(v)=\infty$ for absent candidates. The final ranking uses
\begin{equation}
    S_t(v)=\frac{1}{\pi_z(v)}
        +\frac{\mathbf{1}[j_t\in\mathcal H_t]}{\pi_C(v)}.
    \label{eq:rank_fusion}
\end{equation}
Ties follow $\pi_z(v)$. Local sources contribute historical ranks even when their tokens differ from $r_t$; bank sources use state-based ranks.

\paragraph{Root Copy-row update.}
Up to four ranked candidates refresh the root's eight-entry Copy row $\mathcal U_t$. A target-generated row preserves its original top-one prediction, inserts distinct proposals, then fills remaining slots from the original row. If every proposal is already present, the order stays unchanged.

Copy rows start with up to two offline prior tokens and are refreshed by target predictions during prefill and verification. The updated root row guides drafting. Section~\ref{sec:supp_recap_reranking} evaluates this readout with RECAP's contextual prediction memory and best-first tree policy~\cite{zhu2026recap}.

\subsection{Contextual Draft-Tree Construction}
\label{sec:supp_tree_construction}

\paragraph{Text continuation retrieval.}
The refreshed root row $\mathcal U_t$ carries the state-based proposals into both root edges and continuation retrieval. Successor-guided queries follow LogitSpec~\cite{liu2026logitspec}; here the reranked successors determine which shifted keys are queried. The request-local text index stores token spans from the prompt and accepted output. Let $\oplus$ denote token concatenation and $\operatorname{suf}_m$ the final $m$ tokens. The ordinary and shifted retrieval keys are
\begin{equation}
    \begin{aligned}
        k^{(0)}_{t,m}
            &= \operatorname{suf}_m(P_t\oplus r_t),\\
        k^{(1)}_{t,m}(v)
            &= \operatorname{suf}_m(P_t\oplus(r_t,v)),
            \quad v\in\mathcal U_t.
    \end{aligned}
    \label{eq:text_queries}
\end{equation}
Ordinary retrieval pools matching continuations over available suffix lengths $m\in\mathcal M=\{2,\ldots,10\}$. Shifted retrieval extends the matching endpoint by one proposed successor. For each eligible $v$, it takes the first matching shifted key reached from the trie cursor and follows the most frequent child at each subsequent step.

Let $\mathcal R_t(k)$ denote continuations retrieved after key $k$, and let $c_t(v)$ be the shifted continuation following $v$, or $\epsilon$ when unavailable. The resulting root-aligned paths are
\begin{equation}
    \begin{aligned}
        \mathcal C_t^{(0)}
            &= \bigcup_{m\in\mathcal M}
               \{r_t\oplus c:
                 c\in\mathcal R_t(k^{(0)}_{t,m})\},\\
        \mathcal C_t^{(1)}
            &= \{(r_t,v)\oplus c_t(v):
                 v\in\mathcal U_t,\ c_t(v)\ne\epsilon\}.
    \end{aligned}
    \label{eq:retrieved_paths}
\end{equation}
We retain ordinary continuations by frequency, then insert shifted paths in root-row order. Shared prefixes occupy one node; Copy expansion fills the remaining budget.

\paragraph{Copy-based tree expansion.}
At deeper nodes, ReCue reuses cached target predictions indexed by the parent--current token pair $(a,b)$. A missing pair-specific row falls back to the row indexed by $b$ alone.

Let $\alpha_i$ and $\beta_i$ be calibrated rank weights for root and deeper edges, respectively, with $\beta_\Sigma=\sum_{j=1}^8\beta_j$. An unknown row receives weight $\eta$. For a token pair $(a,b)$ previously scored during target verification, the cached logits $\boldsymbol{\xi}_{a,b}$ define
\[
    p_{a,b,i}
        =\operatorname{softmax}(\boldsymbol{\xi}_{a,b})_i.
\]
These probabilities redistribute the calibrated mass of the corresponding deeper Copy row. The edge weight and path priority are
\begin{equation}
    \begin{aligned}
        \omega_i(n)&=
        \begin{cases}
            \alpha_i, & n\text{ is the root},\\
            \beta_\Sigma p_{a,b,i},
                & \text{scored pair},\ b\ne r_t,\\
            \beta_i, & \text{populated row},\\
            \eta, & \text{otherwise},
        \end{cases}\\
        F(\pi)&=\sum_{(n,i)\in\pi}\log\omega_i(n).
    \end{aligned}
    \label{eq:copy_path_score}
\end{equation}
Here $(n,i)$ identifies the edge from node $n$ to its $i$th cached successor along a Copy path $\pi$. ReCue expands the highest-priority frontier edge until reaching the verification budget of $N=64$ nodes, including the root. If the scored frontier is exhausted before the budget is filled, rank-based Copy expansion supplies the remaining nodes.

We fit rank weights once per target on verified output prefixes from the calibration trajectories. Root and populated Copy rows are counted separately over eight ranks and an outside-list outcome. Each rank receives $\frac{c_i+1}{n_c+9}$, where $c_i$ is its count and $n_c$ includes all nine outcomes. Unknown rows use $\eta=\frac{c_u+1}{n_u+2}$, with $c_u$ fallback-token matches among $n_u$ observations. These profiles stay fixed during evaluation.

Before verification, the original root row is restored. The resulting tree is verified in parallel using the greedy tree-verification procedure described in the main paper.

\paragraph{Online memory update.}
Verification jointly refreshes Copy rows and inserts eligible text nodes, including rejected branches, into the transition history in tree-node order. Each increment is measured from its parent state, using $q_t$ for the root. Accepted text updates the retrieval index, and the final accepted node supplies $q_{t+1}$ and $r_{t+1}$ for the next decoding round.

\section{State-Estimation Controls}
\label{sec:supp_state_estimation}

Fixed-prefix replay holds the prefix, root, candidate source and pool, rank fusion, tree, and verifier fixed. Both routes share the frozen language head and drafting rules. Complete requests provide primary online MAT and throughput; length-sensitivity controls use a common output cap set to each request's shortest completed generation across policies.

\subsection{Post-Root Estimate Quality}
\label{sec:supp_estimate_quality}

On 88 independent requests per target from CharXiv, TextVQA, and MMMU-Pro, we compare ReCue's root updates with cached Copy rows under the native history and residual routes. Comparisons share target observations and a 64-node budget, with EOS or a 1,024-token cap. The Qwen3 cohort also supplies the readout controls in Sec.~\ref{sec:supp_matched_ranking}. Intervals use 100,000 benchmark-stratified source-image bootstrap draws that retain all three targets, with simultaneous 95\% coverage across the six route/target comparisons.

\begin{table}[!htbp]
\centering\small
\setlength{\tabcolsep}{2pt}
\caption{Accepted tokens per round after root updates.}
\label{tab:supp_state_replay}
\begin{tabular*}{\columnwidth}{@{\extracolsep{\fill}}lrrrr@{}}
\toprule
Route & Share (\%) & Cached & ReCue & Gain [CI] \\
\midrule
\multicolumn{5}{@{}l}{\textit{Qwen2.5-VL-7B}}\\
History & 65.7 & 4.02 & 4.22 & $+0.20\;[0.14,0.24]$ \\
Residual & 34.3 & 2.34 & 2.90 & $+0.56\;[0.44,0.68]$ \\
\midrule
\multicolumn{5}{@{}l}{\textit{Qwen3-VL-8B}}\\
History & 61.0 & 3.74 & 3.94 & $+0.20\;[0.17,0.23]$ \\
Residual & 39.0 & 2.32 & 2.78 & $+0.46\;[0.40,0.53]$ \\
\midrule
\multicolumn{5}{@{}l}{\textit{LLaVA-NeXT Vicuna-7B}}\\
History & 69.9 & 4.43 & 4.60 & $+0.17\;[0.12,0.23]$ \\
Residual & 30.1 & 2.22 & 2.71 & $+0.49\;[0.37,0.61]$ \\
\bottomrule
\end{tabular*}
\end{table}

Both routes improve accepted length across all three targets (Tab.~\ref{tab:supp_state_replay}).

The online comparison uses 88 Qwen3-VL-8B requests in three interleaved repeats with an 8,192-token KV capacity. Controls retain cached root rows on history rounds or all rounds; the former keeps residual proposals active. All variants execute source selection and scoring. Intervals use 30,000 benchmark-stratified whole-request bootstrap draws retaining all repeats, with Bonferroni-corrected simultaneous 95\% coverage. Historical root updates raise complete-request MAT from 3.27 to 3.49 and throughput from 73.2 to 77.9 token/s; 61.0\% of rounds have compatible history (Tab.~\ref{tab:supp_history_root_online}).

\begin{table}[!tbp]
\centering\small
\setlength{\tabcolsep}{2.5pt}
\renewcommand{\arraystretch}{1.03}
\caption{Online gains from root updates. The historical-round control retains residual proposals. Brackets give simultaneous 95\% intervals.}
\label{tab:supp_history_root_online}
\begin{tabular}{lrr}
\toprule
Root updates off & $\Delta$MAT & Decode gain (\%) \\
\midrule
\multicolumn{3}{@{}l}{\textit{Qwen3-VL-8B}}\\
\multicolumn{3}{@{}l}{\textit{Complete requests}}\\
Historical rounds & $+0.22\;[0.07,\,0.46]$ & $+6.5\;[2.3,\,13.8]$ \\
All rounds & $+0.29\;[0.15,\,0.52]$ & $+8.9\;[4.5,\,16.3]$ \\
\addlinespace[3pt]
\multicolumn{3}{@{}l}{\textit{Length-sensitivity control}}\\
Historical rounds & $+0.16\;[0.07,\,0.28]$ & $+4.9\;[2.0,\,8.3]$ \\
All rounds & $+0.22\;[0.13,\,0.29]$ & $+6.8\;[3.8,\,9.1]$ \\
\bottomrule
\end{tabular}
\end{table}

State probes for the three targets use 100 image requests each (20 per main benchmark) and 50 video requests along residual-predictor trajectories. Main-paper Table~5 uses 5,000 benchmark-stratified whole-request bootstrap draws.

We measure transfer error as
\begin{equation}
e_j=\frac{\lVert\Delta_j-(h_t^+-q_t)\rVert_2}
{\max(\lVert h_t^+-q_t\rVert_2,10^{-12})},
\end{equation}
where $h_t^+$ is the true post-root state. Predictor error replaces $\Delta_j$ with $\widehat z_t-q_t$. Relative to residual estimates, historical increments reduce mean relative error by 45--54\% on images and 44--50\% on video. Top-1 successor recall improves by 4.9/5.4/6.3 percentage points on images and 4.8/5.5/4.3 on video, in target order, with positive simultaneous 95\% request-bootstrap intervals for all six gains. Top-1 recall measures whether the first proposal matches the target's post-root prediction on nonterminal rounds. Source similarity correlates with lower transfer error: Spearman $\rho$ ranges from $-0.86$ to $-0.71$ on images and $-0.88$ to $-0.79$ on video.

\begin{figure}[!tbp]
    \centering
    \includegraphics[width=\linewidth]{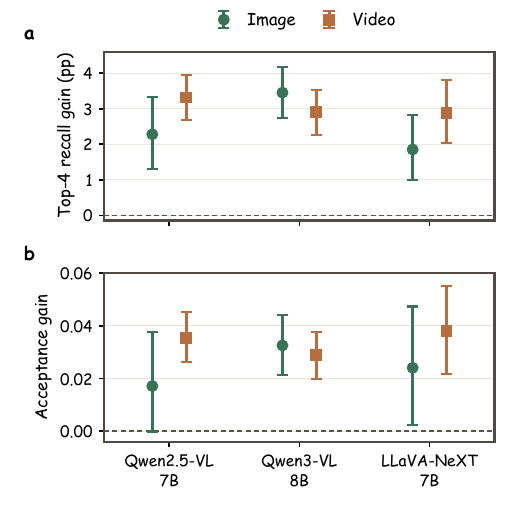}
    \caption{Increment transfer versus direct state reuse at fixed prefixes. Top: Top-4 recall gain. Bottom: accepted-token gain per round. Intervals are simultaneous 95\% bounds.}
    \label{fig:supp_transfer_error}
\end{figure}

Applying the historical increment to $q_t$ raises Top-4 successor recall by 1.9--3.5 percentage points over directly reusing $h_{j_t}^+$ across the three targets and both modalities (Fig.~\ref{fig:supp_transfer_error}). Other rounds share the residual estimate. On Qwen3-VL, the paired accepted-length gain is $+0.03\;[+0.02,+0.04]$ tokens per round for both images and video. Intervals use 100,000 benchmark-stratified whole-request bootstrap draws with simultaneous 95\% coverage across the six target/modality contrasts.

\paragraph{Source selection.}
The comparison uses 100 independent requests per target from CharXiv, TextVQA, and MMMU-Pro, a 64-node budget, and shared residual estimates on unmatched rounds. Benchmark-stratified whole-request bootstraps provide simultaneous 95\% intervals across both targets. ReCue's source-selection rule reduces transition error by 4.0\% and 2.6\% relative to averaging same-root increments, and by 10.0\% and 6.7\% relative to using the latest increment, for Qwen2.5-VL and Qwen3-VL, respectively. Top-1 recall gains over mean increments are $+0.88\;[+0.06,+1.74]$ and $+0.83\;[+0.28,+1.35]$ percentage points, versus 4.7 and 5.3 points over residual estimates.

\subsection{Candidate-Ranking Controls}
\label{sec:supp_matched_ranking}
We compare four readouts on 88 independent Qwen3-VL requests from CharXiv, TextVQA, and MMMU-Pro along native ReCue prefixes, sharing cache observations and a 64-node budget. For compatible local records, ReCue applies the historical increment to the current state; alternatives use the historical post-root state, cached successor order, or linear logit fusion. Other rounds share the residual estimate. Recall uses nonterminal roots. Simultaneous 95\% intervals across the three controls use 100,000 benchmark-stratified whole-request bootstrap draws. Logit fusion uses
\begin{equation}
\ell_{\alpha}(v)=(1-\alpha)(Wq_t)_v+\alpha(Wh_{j_t}^{+})_v.
\end{equation}
Here $\alpha\in\{0.25,0.50,0.75\}$; the fixed-prefix comparison uses $\alpha=0.50$.

\begin{table}[!tbp]
\centering\small
\setlength{\tabcolsep}{4pt}
\caption{Qwen3-VL-8B readouts at fixed prefixes: accepted length $\overline A$ and Top-4 recall (\%).}
\label{tab:supp_matched_ranking_replay}
\begin{tabular*}{\columnwidth}{@{\extracolsep{\fill}}lrr@{}}
\toprule
Readout & $\overline A$ & Top-4 \\
\midrule
Historical state & 3.45 & 67.3 \\
Cached successor order & 3.45 & 67.3 \\
Logit fusion ($\alpha=0.50$) & 3.46 & 67.8 \\
\textbf{ReCue} & \textbf{3.49} & \textbf{70.7} \\
\bottomrule
\end{tabular*}
\end{table}

ReCue raises Top-4 successor recall from 67.3\% to 70.7\% and accepted length from 3.45 to 3.49 relative to direct historical-state reuse (Tab.~\ref{tab:supp_matched_ranking_replay}). The paired accepted-length gain is $+0.04\;[+0.02,+0.05]$ tokens per round. ReCue also increases accepted length relative to cached successor order and equal-weight logit fusion.

\subsection{Multi-Turn Candidate Reranking}
\label{sec:supp_multiturn_reranking}
Recall uses fixed ReCue prefixes and natural candidate pools. Online runs use complete requests and the common-cap control defined above.

\paragraph{Official conversations.}
We evaluate Qwen3-VL-8B on the 223 original questions in 100 MMSpec conversation sessions~\cite{shen2026mmspec}. All variants receive the same archived LLaVA AR answers as earlier-turn context. The source-order control inserts the selected source's cached successors into the root row. A second control retains cached root rows on history rounds and keeps residual proposals active. Intervals use 30,000 whole-conversation bootstrap draws, with simultaneous 95\% coverage across 84 recall and MAT comparisons spanning targets, output settings, and all-turn/follow-up groups.

ReCue improves Top-4 recall by 4.6 percentage points on history rounds and 11.3 points on residual rounds. MAT rises from 2.87 to 2.99 on complete requests and from 2.87 to 2.96 with a common output cap (Tab.~\ref{tab:supp_official_conversation}). Historical root updates add $+0.11\;[0.01,\,0.22]$ tokens per round on complete requests and $+0.09\;[0.04,\,0.14]$ under the common cap.

\begin{table}[!tbp]
\centering\small
\setlength{\tabcolsep}{1.5pt}
\renewcommand{\arraystretch}{1.05}
\caption{Qwen3-VL-8B on MMSpec conversations.}
\label{tab:supp_official_conversation}
\begin{tabular*}{\columnwidth}{@{\extracolsep{\fill}}lrrr@{}}
\toprule
Route & Source order & ReCue & $\Delta$Top-4 (pp) \\
\midrule
History & 74.6 & 79.2 & $+4.6\;[4.0,\,5.2]$ \\
Residual & 30.7 & 42.1 & $+11.3\;[9.7,\,13.1]$ \\
\bottomrule
\end{tabular*}
\vspace{3pt}
\begin{tabular*}{\columnwidth}{@{\extracolsep{\fill}}lrrr@{}}
\toprule
Root policy & Control MAT & ReCue & $\Delta$MAT \\
\midrule
\multicolumn{4}{@{}l}{\textit{Complete requests}}\\
Source order & 2.87 & 2.99 & $+0.12\;[0.03,\,0.24]$ \\
Cached history roots & 2.88 & 2.99 & $+0.11\;[0.01,\,0.22]$ \\
\addlinespace[2pt]
\multicolumn{4}{@{}l}{\textit{Common output cap}}\\
Source order & 2.87 & 2.96 & $+0.09\;[0.02,\,0.18]$ \\
Cached history roots & 2.87 & 2.96 & $+0.09\;[0.04,\,0.14]$ \\
\bottomrule
\end{tabular*}
\end{table}

\paragraph{AVSD video dialogues.}
We evaluate Qwen3-VL-8B on 32 AVSD validation videos with all ten original turns per dialogue (320 questions)~\cite{alamri2019avsd}. All policies use up to 32 video frames, a 16,384-token KV capacity, and the same earlier-turn answers generated by full ReCue. Intervals use 30,000 whole-video bootstrap draws with simultaneous 95\% coverage across 32 contrasts. Relative to source order, Top-4 recall improves by $5.7\;[4.6,\,6.8]$ percentage points on history rounds and $15.2\;[12.9,\,17.7]$ on residual rounds. Complete-request online MAT rises from 2.65 to 2.72, a gain of $+0.07\;[0.03,\,0.12]$ tokens per round. Historical root updates add $+0.09\;[0.05,\,0.13]$ over cached roots on history rounds, with residual proposals active in both policies.

\paragraph{Three-turn video questions.}
We evaluate Qwen3-VL-8B on 12 LongVideoBench and 12 Video-MME videos with three follow-up questions about the opening, subsequent changes, and ending. We construct these questions and share the earlier-turn transcript from full ReCue across policies, using native video inputs with up to 32 frames. The source-order control shares the tree and verification rules. Intervals use 30,000 benchmark-stratified whole-video bootstrap draws retaining all three turns. Simultaneous 95\% coverage spans 108 contrasts across targets, root policies, recall and MAT, output settings, and pooled and per-benchmark results.

ReCue improves Top-4 recall by 4.4 percentage points on history rounds and 12.0 points on residual rounds. Complete-request MAT rises from 2.89 to 3.04, a gain of $+0.15\;[0.05,\,0.35]$ tokens per round. With a common output cap, MAT rises from 2.65 to 2.78 on LongVideoBench and from 3.09 to 3.23 on Video-MME (Tab.~\ref{tab:supp_multiturn_reranking}).

\begin{table}[!tbp]
\centering\small
\setlength{\tabcolsep}{1.5pt}
\renewcommand{\arraystretch}{1.05}
\caption{Qwen3-VL-8B on three-turn video questions.}
\label{tab:supp_multiturn_reranking}
\begin{tabular*}{\columnwidth}{@{\extracolsep{\fill}}lrrr@{}}
\toprule
Route & Source order & ReCue & $\Delta$Top-4 (pp) \\
\midrule
History & 73.8 & 78.2 & $+4.4\;[2.8,\,6.1]$ \\
Residual & 35.5 & 47.4 & $+12.0\;[9.7,\,14.3]$ \\
\bottomrule
\end{tabular*}
\vspace{3pt}
\begin{tabular*}{\columnwidth}{@{\extracolsep{\fill}}lrrr@{}}
\toprule
\multicolumn{4}{@{}l}{\textit{Online MAT: common output cap}}\\
Workload & Source order & ReCue & $\Delta$MAT \\
\midrule
Pooled & 2.89 & 3.02 & $+0.13\;[0.06,\,0.21]$ \\
LongVideoBench & 2.65 & 2.78 & $+0.13\;[0.05,\,0.20]$ \\
Video-MME & 3.09 & 3.23 & $+0.13\;[0.03,\,0.27]$ \\
\bottomrule
\end{tabular*}
\end{table}

\subsection{Reranking on the RECAP Tree Policy}
\label{sec:supp_recap_reranking}
We apply ReCue's state-based candidate readout while retaining RECAP's compiled context-memory lookup, confidence fusion, deeper expansion, and suffix policy. Up to four proposals enter the root row. Insertion renormalizes native rank-probability masses, geometrically decays masses for added slots, and initializes empty rows by reciprocal rank. All arms execute 64 positions, padding unused slots with unobserved leaves. We compare native RECAP, cache-only maintenance without source queries, historical ranking, and full reranking. Native and cache-only outputs and paths match on every request. Each Qwen target uses 100 requests, 20 each from GQA, TextVQA, COCO, CharXiv, and MMMU-Pro, with natural EOS or 256 tokens and three interleaved repeats. Persistent memory resets for each of ten 10-request streams, two per benchmark. Paired intervals resample benchmark-stratified 10-request streams, retaining all three repeats.

\begin{table}[!tbp]
\centering\small
\setlength{\tabcolsep}{2.5pt}
\renewcommand{\arraystretch}{1.03}
\caption{RECAP integration. ReCue gain is decode throughput relative to each control (\%); $^{*}$99.75\% intervals, otherwise 95\%.}
\label{tab:supp_recap_reranking}
\begin{tabular}{p{79pt}rrr}
\toprule
Policy & MAT & tok/s & ReCue gain \\
\midrule
\multicolumn{4}{@{}l}{\textit{Qwen2.5-VL-7B}}\\
RECAP$^{*}$ & 2.66 & 75.2 & $+9.0\;[7.1,\,11.1]$ \\
+ cache only & 2.66 & 73.4 & $+11.7\;[10.3,\,13.1]$ \\
+ historical ranking & 2.94 & 79.8 & $+2.8\;[1.3,\,4.3]$ \\
\textbf{+ ReCue reranking} & \textbf{3.03} & \textbf{82.0} & -- \\
\addlinespace[3pt]
\multicolumn{4}{@{}l}{\textit{Qwen3-VL-8B}}\\
RECAP$^{*}$ & 2.61 & 59.0 & $+9.3\;[7.8,\,10.9]$ \\
+ cache only & 2.61 & 57.8 & $+11.7\;[10.3,\,13.0]$ \\
+ historical ranking & 2.88 & 62.9 & $+2.6\;[1.5,\,3.6]$ \\
\textbf{+ ReCue reranking} & \textbf{2.97} & \textbf{64.5} & -- \\
\bottomrule
\end{tabular}
\end{table}

On Qwen2.5-VL, reranking raises MAT from 2.66 to 3.03 and decode throughput from 75.2 to 82.0 token/s; on Qwen3-VL, MAT rises from 2.61 to 2.97 and throughput from 59.0 to 64.5 token/s. The paired decode throughput gains over native RECAP are $+9.0 [7.1, 11.1]\%$ and $+9.3 [7.8, 10.9]\%$, respectively, on Qwen2.5-VL and Qwen3-VL (99.75\% paired intervals). Request throughput increases by $7.2\%$ and $8.8\%$, respectively.

Cache maintenance adds $+2.5 [2.1, 2.8]\%$ and $+2.1 [2.0, 2.3]\%$ to decode time under identical generated tokens and accepted paths (95\% paired stream intervals).

\subsection{Online State-Estimation Policies}

On Qwen2.5-VL-7B and Qwen3-VL-8B, we compare ReCue's estimator with a variant that uses the residual predictor on every round. Both variants share the trained head and the rules for source selection, candidate-pool construction, rank fusion, and 64-node tree construction. Each target uses 50 LongVideoBench and 50 Video-MME requests, three interleaved repeats, and natural EOS with a 1,024-token cap. Paired video bootstraps retain both targets and all repeats, with simultaneous 95\% coverage across workloads within each metric.

\begin{figure}[!tbp]
    \centering
    \includegraphics[width=\linewidth]{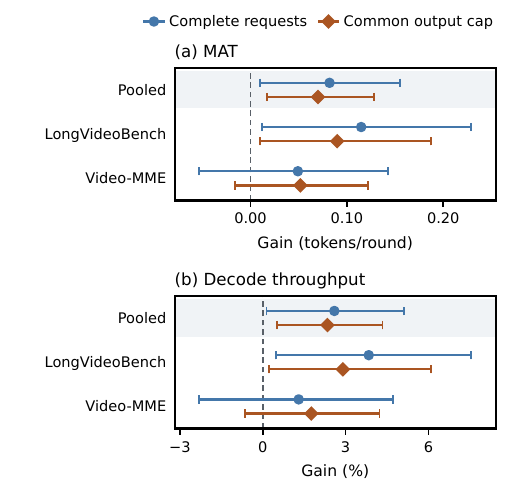}
    \caption{Estimator-policy gains on complete requests and with a common output cap. Intervals are simultaneous 95\% bounds across workloads within each metric and condition.}
    \label{fig:supp_history_video_mean}
\end{figure}

Across the two targets, mean MAT increases by $0.08\;[0.01,0.16]$ tokens per round, and the geometric mean of target-level decode-throughput ratios improves by $2.6\%\;[0.1,5.1]$ overall and $3.8\%\;[0.5,7.5]$ on LongVideoBench (Fig.~\ref{fig:supp_history_video_mean}). Under the common output cap, the gains are $+0.07\;[+0.02,+0.13]$ tokens per round in MAT and $+2.3\%\;[+0.5,+4.3]$ in decode throughput.

\section{Visual Conditioning Experiments}
\label{sec:supp_visual_evidence}

\subsection{Controlled Visual Changes}

We test the residual branch by changing one legend digit under fixed text. The main-paper control uses 40 templates, each with ten digit variants (400 charts per target), spanning five legend placements, two fonts, and four curve/axis styles. Each template fixes the prompt, forced root \texttt{1}, source, candidate IDs, and geometry; the first variant supplies the source, pool, and frozen ranking. The pool contains all ten candidate digits. Pixel checks localize changes to the edited digit.

\begin{table}[!htbp]
\centering\small
\setlength{\tabcolsep}{2.5pt}
\renewcommand{\arraystretch}{1.03}
\caption{Legend intervention on 400 charts per target. Top-$k$ coverage (\%) uses 95\% template-bootstrap intervals; the first two rows count cases.}
\label{tab:supp_controlled_400}
\begin{tabular*}{\columnwidth}{@{\extracolsep{\fill}}lrr@{}}
\toprule
Measurement & Qwen2.5-VL & Qwen3-VL \\
\midrule
Natural root 1 & 400/400 & 397/400 \\
Teacher successor & 400/400 & 400/400 \\
Frozen top-4 (\%) & 40.00 & 40.00 \\
ReCue top-1 (\%) [CI] & \shortstack[r]{92.75\\{[89.50, 95.75]}} & \shortstack[r]{24.75\\{[20.00, 29.75]}} \\
ReCue top-4 (\%) [CI] & \shortstack[r]{100.00\\{[100.00, 100.00]}} & \shortstack[r]{80.00\\{[76.25, 83.50]}} \\
\bottomrule
\end{tabular*}
\end{table}

ReCue's top-4 coverage reaches 100\% and 80\% for Qwen2.5 and Qwen3, versus 40\% for the frozen ranking (Tab.~\ref{tab:supp_controlled_400}). The natural root is \texttt{1} on 400/400 and 397/400 cases. After the forced root, the target predicts the edited successor in all 400 cases per target. Bootstrap draws resample whole templates with their ten variants.

\paragraph{Two-layout control.}
The two-layout control uses red-curve labels from 10,000 to 19,000. Within each layout, the second legend digit changes while curves, axes, other labels, question, and dimensions stay fixed. Pixel checks localize changes to the legend entry. Both targets naturally select root \texttt{1} and predict its successor correctly in all 20 cases. Each layout's first image supplies the source, pool, and frozen ranking for the other nine. The pool contains all ten successors, and the frozen residual predictor supplies the proposal state.

\begin{table}[!htbp]
    \centering
    \caption{Correct-successor coverage (\%) on 20 controlled charts per target. Top-4 counts four distinct proposals.}
    \label{tab:supp_controlled_visual}
    \small
    \begin{tabular*}{\columnwidth}{@{\extracolsep{\fill}}lrrrr@{}}
        \toprule
        & \multicolumn{2}{c}{Frozen ranking} & \multicolumn{2}{c}{Current visual state} \\
        \cmidrule(lr){2-3}\cmidrule(lr){4-5}
        Target & Top-1 & Top-4 & Top-1 & Top-4 \\
        \midrule
        Qwen2.5-VL-7B & 10 & 40 & 100 & 100 \\
        Qwen3-VL-8B & 10 & 40 & 15 & 80 \\
        \bottomrule
    \end{tabular*}
\end{table}

\subsection{Natural-Image Query Interventions}

Main-paper Table~6 uses 100 images per Qwen target with a 256-token cap. Replay fixes the accepted prefix, root, source, increment, candidate IDs, Copy observations, retrieved continuations, and original-image KV prefix. The historical-ordering contrast changes ranking; the image-query contrast replaces the current query used for post-root scoring. Both keep the original visually conditioned memories and verifier.

We further evaluate both Qwen targets on the same 100 held-out COCO images, paired into 50 reciprocal donor pairs. Each request asks, ``Describe what is visible in this image in one concise sentence.'' Generation stops at EOS or 1,024 tokens. Donor images preserve the recipient's dimensions and visual-token geometry. Fixed-prefix replay changes only the current query; source, increment, candidate pool, tree policy, and original-image verifier stay fixed. Online arms follow their own accepted text and histories, with donor states computed along the same text as each arm. Self-image controls reproduce native outputs and accepted paths.

\begin{figure}[!tbp]
    \centering
    \includegraphics[width=\linewidth]{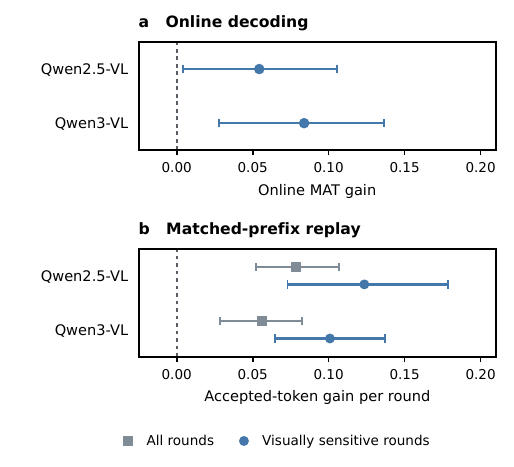}
    \caption{Original versus donor queries in COCO captioning. MAT intervals are 97.5\%; replay intervals are 95\%.}
    \label{fig:supp_caption_query}
\end{figure}

Original-image queries increase online MAT by $+0.054\;[+0.004,+0.106]$ and $+0.084\;[+0.028,+0.136]$ on Qwen2.5-VL and Qwen3-VL, respectively (Fig.~\ref{fig:supp_caption_query}). Intervals use 20,000 whole-pair bootstrap draws, with 97.5\% bounds for the two online contrasts.

\begin{figure}[!tbp]
    \centering
    \includegraphics[width=\linewidth]{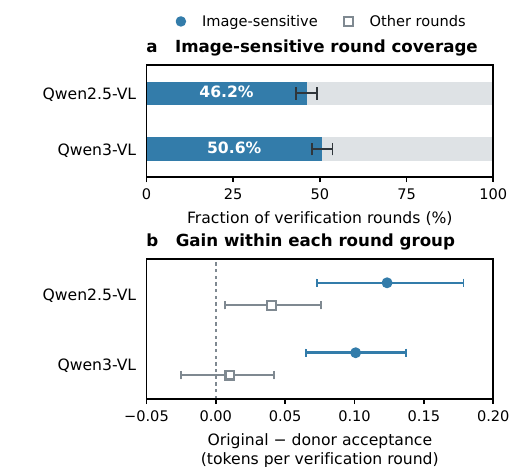}
    \caption{Query-image effects on COCO. Sensitive rounds change the target successor under the donor image. Intervals are 95\% donor-pair bootstrap bounds.}
    \label{fig:supp_visual_dependence}
\end{figure}

With text and root fixed, substituting the donor image changes the target's successor prediction in 46.2\% of verification rounds for Qwen2.5-VL and 50.6\% for Qwen3-VL. On these image-sensitive rounds, the original query improves replay acceptance by 0.123 and 0.101 tokens per round, compared with 0.040 and 0.010 on the remaining rounds. The gaps between the two groups are $+0.084\;[+0.007,+0.165]$ and $+0.091\;[+0.039,+0.144]$ tokens per round, respectively, with simultaneous 95\% donor-pair bootstrap bounds across the two targets (Fig.~\ref{fig:supp_visual_dependence}).

\subsection{Successor Rankings and Accepted Paths}

\begin{figure}[!tbp]
    \centering
    \includegraphics[width=\linewidth]{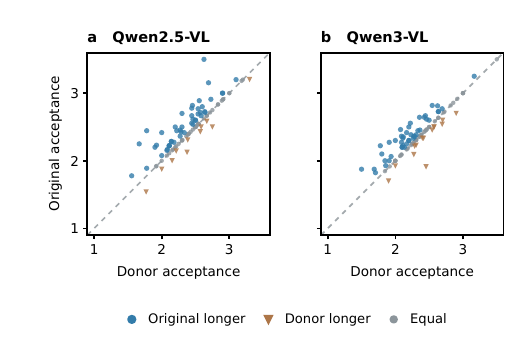}
    \caption{Per-request mean accepted tokens per round on COCO. Dashed line: equality.}
    \label{fig:supp_all_visual_images}
\end{figure}

Using the same COCO replay protocol on 100 requests per target, original queries yield longer accepted paths in 68/76 rounds and shorter paths in 23/20 across 1,070/1,195 verification rounds on Qwen2.5-VL/Qwen3-VL. The net gains are 84/67 accepted tokens (Tab.~\ref{tab:supp_caption_paths}). Figure~\ref{fig:supp_all_visual_images} shows each request's mean accepted tokens per round.

\begin{table}[!tbp]
\centering\footnotesize
\setlength{\tabcolsep}{2.5pt}
\caption{Fixed-prefix acceptance gains and losses on COCO.}
\label{tab:supp_caption_paths}
\begin{tabular}{lrrrl}
\toprule
Target & \shortstack{Original\\longer} & \shortstack{Donor\\longer} & \shortstack{Net\\tokens} & $\Delta A$ [95\% CI] \\
\midrule
Qwen2.5-VL & 68 & 23 & +84 & $+0.079\;[+0.052,+0.107]$ \\
Qwen3-VL & 76 & 20 & +67 & $+0.056\;[+0.029,+0.083]$ \\
\bottomrule
\end{tabular}
\end{table}

\begin{figure*}[!tbp]
    \centering
    \includegraphics[width=\textwidth]{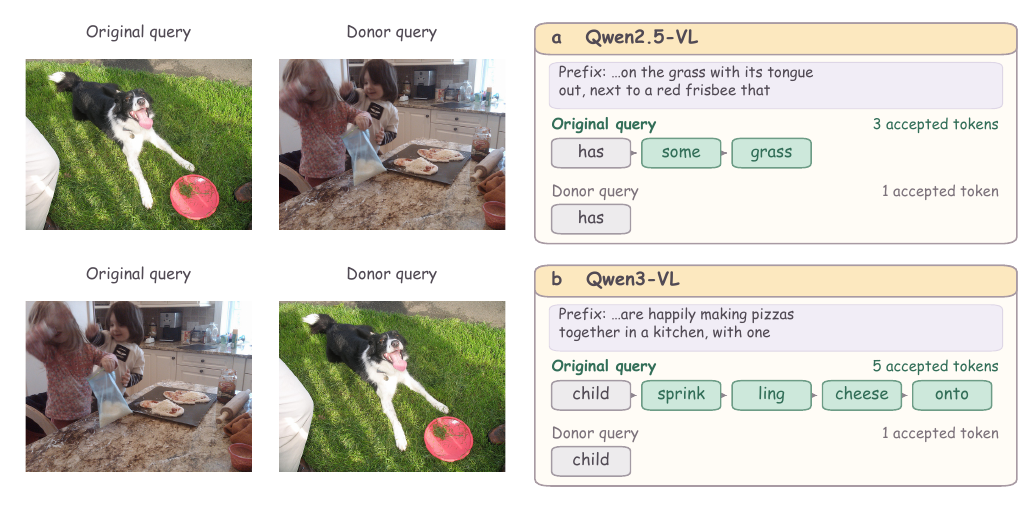}
    \caption{Original and donor query examples. Gray marks the fixed root; green marks additional accepted tokens.}
    \label{fig:supp_visual_tree}
\end{figure*}

Figure~\ref{fig:supp_visual_tree} illustrates a reciprocal donor pair. The original query extends acceptance to \texttt{has some grass} (three tokens) for the frisbee scene and \texttt{child sprinkling cheese onto} (five tokens) for the pizza scene. Acceptance stops at the fixed root for both the donor query and predictor-only control.

\begin{figure*}[!tbp]
    \centering
    \includegraphics[width=\textwidth]{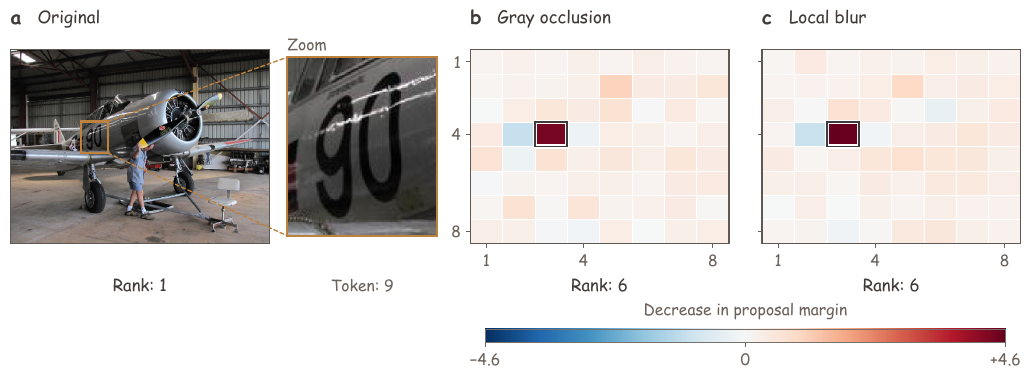}
    \caption{Aircraft-number reading with Qwen2.5-VL-7B. Outlined map cells mark the largest margin decrease.}
    \label{fig:supp_number_regions}
\end{figure*}

\paragraph{Spatial interventions.}
For the aircraft-number probe, we replace one cell at a time in an $8\times8$ grid with RGB $(128,128,128)$ or its Gaussian-blurred crop (radius 16 original-image pixels). Image size, text, visual-token geometry, source, increment, pool, and root stay fixed. For target successor $v$, the proposal margin is
\begin{equation}
    m_t(v)=(Wz_t)_v-\max_{u\in\mathcal V_t\setminus\{v\}}(Wz_t)_u.
\end{equation}
Maps show the intervention-induced decrease in margin, using unsmoothed signed responses from all 64 cells and a shared scale. Peaks coincide with the aircraft number (Fig.~\ref{fig:supp_number_regions}). For the first digit of aircraft number \texttt{90}, rank changes from 1 to 6 under either regional intervention and to 29 under the donor image. Both aircraft queries retain three accepted tokens.

\section{System Measurements}
\label{sec:supp_system}

\subsection{Request Latency and Output Length}
ReCue reduces mean end-to-end latency in all six SGLang target--modality comparisons (Tab.~\ref{tab:supp_sglang_latency}). On Qwen3-VL video, mean latency falls from 19.201\,s to 11.242\,s.

\begin{table}[!htbp]
\centering
\small
\setlength{\tabcolsep}{3pt}
\renewcommand{\arraystretch}{1.07}
\caption{Single-request SGLang latency and decode throughput. E2E is mean end-to-end seconds per request.}
\label{tab:supp_sglang_latency}
\begin{tabular*}{\columnwidth}{@{\extracolsep{\fill}}llrrrr@{}}
\toprule
& & \multicolumn{2}{c}{Decode token/s} & \multicolumn{2}{c}{E2E (s)} \\
\cmidrule(lr){3-4}\cmidrule(lr){5-6}
Target & Input & AR & ReCue & AR & ReCue \\
\midrule
Qwen2.5-VL-7B & Image & 33.77 & 93.85 & 4.384 & 1.711 \\
Qwen3-VL-8B & Image & 31.75 & 78.56 & 10.296 & 4.246 \\
LLaVA-NeXT-7B & Image & 32.15 & 79.90 & 4.944 & 2.172 \\
Qwen2.5-VL-7B & Video & 32.55 & 62.03 & 12.640 & 7.710 \\
Qwen3-VL-8B & Video & 30.72 & 65.60 & 19.201 & 11.242 \\
LLaVA-NeXT-7B & Video & 31.53 & 73.36 & 15.262 & 7.162 \\
\bottomrule
\end{tabular*}
\end{table}

Across the three image targets, longer AR-output bins have higher MAT and speedup, reaching $2.81$--$3.47\times$ AR throughput above 256 tokens (Tab.~\ref{tab:supp_output_length}). We group native main-table requests by AR output length and keep method pairs in the same bin. Each bin pools emitted tokens and measured decode times, including capped outputs.

\begin{table}[!htbp]
\centering\small
\setlength{\tabcolsep}{2.5pt}
\renewcommand{\arraystretch}{1.03}
\caption{Native results grouped by paired AR output length. Throughput pools emitted tokens and decode times within each bin.}
\label{tab:supp_output_length}
\begin{tabular*}{\columnwidth}{@{\extracolsep{\fill}}llrrrr@{}}
\toprule
Input & AR length & $n$ & MAT & tok/s & $\times$AR \\
\midrule
\multicolumn{6}{@{}l}{\textit{Qwen2.5-VL-7B}}\\
Image & $\leq64$ & 233 & 2.972 & 83.54 & 2.50\texttimes \\
Image & 65--256 & 197 & 3.160 & 89.24 & 2.73\texttimes \\
Image & $>256$ & 70 & 3.996 & 110.74 & 3.47\texttimes \\
Video & 65--256 & 46 & 2.647 & 58.95 & 2.15\texttimes \\
Video & $>256$ & 225 & 3.081 & 54.27 & 2.38\texttimes \\
\addlinespace[3pt]
\multicolumn{6}{@{}l}{\textit{Qwen3-VL-8B}}\\
Image & $\leq64$ & 60 & 2.682 & 60.78 & 2.16\texttimes \\
Image & 65--256 & 260 & 2.932 & 65.96 & 2.37\texttimes \\
Image & $>256$ & 180 & 3.453 & 76.15 & 2.81\texttimes \\
Video & 65--256 & 23 & 2.606 & 49.23 & 2.12\texttimes \\
Video & $>256$ & 248 & 3.040 & 38.36 & 2.58\texttimes \\
\addlinespace[3pt]
\multicolumn{6}{@{}l}{\textit{LLaVA-NeXT-7B}}\\
Image & $\leq64$ & 217 & 2.931 & 63.12 & 2.11\texttimes \\
Image & 65--256 & 211 & 3.278 & 72.21 & 2.44\texttimes \\
Image & $>256$ & 72 & 4.635 & 98.21 & 3.46\texttimes \\
Video & $>256$ & 271 & 3.332 & 66.87 & 2.48\texttimes \\
\bottomrule
\end{tabular*}
\end{table}

\subsection{Concurrent Image Serving}
\label{sec:supp_concurrent_serving}

We compare AR, RACER, and ReCue on the same 500 Qwen3-VL-8B requests at 2, 4, and 8 clients. All runs use FlashInfer, decode CUDA Graphs, a 49,152-token KV pool, and 16 server slots, with scheduling overlap disabled. Both speculative methods verify eight positions. Each pass pairs methods on one device; the eight-client setting has three passes with rotated order. Throughput and P95 latency include visual processing and prefill.

\begin{table}[!htbp]
\centering\small
\setlength{\tabcolsep}{2pt}
\renewcommand{\arraystretch}{1.03}
\caption{Concurrent serving on Qwen3-VL-8B.}
\label{tab:supp_serving_a40}
\begin{tabular*}{\columnwidth}{@{\extracolsep{\fill}}clrrrr@{}}
\toprule
Load & Method & MAT & Tok/s & $\times$AR & P95 (s) \\
\midrule
\multirow{3}{*}{2} & AR & 1.00 & 64.0 & 1.00\texttimes & 31.7 \\
 & RACER & 1.60 & 84.8 & 1.32\texttimes & 21.0 \\
 & ReCue & \textbf{1.85} & \textbf{102.5} & \textbf{1.60\texttimes} & \textbf{18.0} \\
\addlinespace[2pt]
\multirow{3}{*}{4} & AR & 1.00 & 123.2 & 1.00\texttimes & 32.7 \\
 & RACER & 1.59 & 142.5 & 1.16\texttimes & 24.5 \\
 & ReCue & \textbf{1.85} & \textbf{180.1} & \textbf{1.46\texttimes} & \textbf{20.1} \\
\addlinespace[2pt]
\multirow{3}{*}{8} & AR & 1.00 & 230.2 & 1.00\texttimes & 34.8 \\
 & RACER & 1.59 & 218.9 & 0.95\texttimes & 31.6 \\
 & ReCue & \textbf{1.83} & \textbf{296.5} & \textbf{1.29\texttimes} & \textbf{24.0} \\
\bottomrule
\end{tabular*}
\end{table}

\begin{figure}[t]
    \centering
    \includegraphics[width=\linewidth]{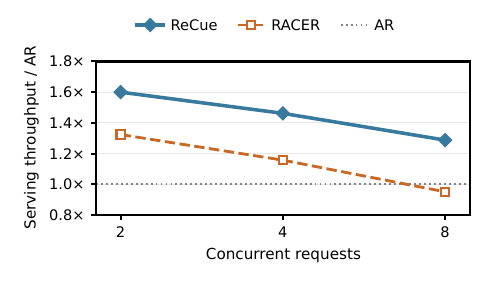}
    \caption{Qwen3-VL-8B throughput relative to AR.}
    \label{fig:recue_concurrent_serving}
\end{figure}

ReCue reaches 1.29$--$1.60$\times$ AR and 1.21$--$1.35$\times$ RACER throughput (Fig.~\ref{fig:recue_concurrent_serving}). Across three eight-client passes, the gain over RACER is 35.2\%--37.5\%.

\subsection{Repeated Native Timing}

\begin{table*}[!htbp]
\centering
\small
\setlength{\tabcolsep}{4pt}
\renewcommand{\arraystretch}{1.07}
\caption{Repeated native timing. Decode and request throughput pool tokens and corresponding times; brackets are 95\% whole-stream intervals.}
\label{tab:supp_repeated_timing}
\begin{tabular}{lrrrrr}
\toprule
Method & MAT & Decode token/s [CI] & $\times$AR & Request token/s [CI] & $\times$AR \\
\midrule
\multicolumn{6}{@{}l}{\textit{Qwen2.5-VL-7B}}\\
AR & 1.00 & 32.6 [32.6, 32.6] & 1.00\texttimes & 30.2 [30.1, 30.3] & 1.00\texttimes \\
RACER & 2.27 & 66.2 [65.3, 66.9] & 2.03\texttimes & 56.4 [55.3, 57.2] & 1.87\texttimes \\
RECAP (64) & 2.71 & 76.5 [75.5, 77.4] & 2.35\texttimes & 64.2 [62.9, 65.3] & 2.13\texttimes \\
RECAP (padded 64) & 2.72 & 76.3 [75.4, 77.1] & 2.34\texttimes & 64.1 [63.0, 65.1] & 2.12\texttimes \\
RECAP (native $\leq68$) & 2.73 & 75.3 [74.5, 76.1] & 2.31\texttimes & 63.5 [62.3, 64.4] & 2.10\texttimes \\
ReCue & 3.03 & 85.0 [84.2, 85.8] & 2.61\texttimes & 69.5 [68.2, 70.5] & 2.30\texttimes \\
\midrule
\multicolumn{6}{@{}l}{\textit{Qwen3-VL-8B}}\\
AR & 1.00 & 27.6 [27.4, 27.7] & 1.00\texttimes & 27.0 [26.9, 27.2] & 1.00\texttimes \\
RACER & 2.30 & 53.2 [52.7, 53.7] & 1.93\texttimes & 51.1 [50.6, 51.5] & 1.89\texttimes \\
RECAP (64) & 2.67 & 60.8 [60.2, 61.3] & 2.20\texttimes & 58.2 [57.6, 58.7] & 2.15\texttimes \\
RECAP (padded 64) & 2.67 & 60.1 [59.3, 60.9] & 2.18\texttimes & 57.5 [56.8, 58.3] & 2.13\texttimes \\
RECAP (native $\leq68$) & 2.69 & 60.3 [59.6, 60.9] & 2.18\texttimes & 57.7 [57.1, 58.3] & 2.13\texttimes \\
ReCue & 2.95 & 66.4 [65.9, 66.8] & 2.41\texttimes & 63.0 [62.5, 63.4] & 2.33\texttimes \\
\bottomrule
\end{tabular}
\end{table*}

We repeat 400 frozen requests three times per target, using 20 independent 20-request streams across GQA, TextVQA, COCO, CharXiv, and MMMU-Pro. Comparisons cover AR, RACER, three RECAP variants (64-node budget, 64 executed positions with padding, or up to 68 native positions), and ReCue. RECAP uses the compiled C++ backend, resetting persistent memory per stream and local history per request. Input shapes are warmed and method order rotates across requests and repeats. Runs use BF16 targets, greedy decoding, a 4,096-token KV capacity, EOS or 256 tokens, PyTorch 2.5.1+cu121, and Transformers 4.57.0.

We compute native decode throughput over the decoding phase. For request throughput, we also time preparation, visual encoding, and prefill. The 20,000 benchmark-stratified whole-stream bootstrap draws retain each request's three repeats. The 16 baseline throughput contrasts use 95\% simultaneous intervals with Bonferroni adjustment; absolute throughput uses pointwise 95\% intervals (Tab.~\ref{tab:supp_repeated_gains}). P50/P95 pool all requests and repeats; per-request timing CV is the standard deviation across repeats divided by their mean (Tab.~\ref{tab:supp_repeated_variability}).

Compared with RECAP using a 64-node budget, ReCue increases decode throughput by $11.18\%$ on Qwen2.5-VL and $9.20\%$ on Qwen3-VL, with corresponding request-throughput gains of $8.17\%$ and $8.27\%$.

\begin{table}[!htbp]
\centering\small
\setlength{\tabcolsep}{2.5pt}
\renewcommand{\arraystretch}{1.03}
\caption{Paired ReCue throughput gains (\%) with simultaneous 95\% intervals across 16 contrasts.}
\label{tab:supp_repeated_gains}
\begin{tabular*}{\columnwidth}{@{\extracolsep{\fill}}p{87pt}rr@{}}
\toprule
Baseline & Decode gain [CI] & Request gain [CI] \\
\midrule
\multicolumn{3}{@{}l}{\textit{Qwen2.5-VL-7B}}\\
RACER & \shortstack[r]{+28.43\\{[+26.83, +30.04]}} & \shortstack[r]{+23.17\\{[+21.86, +24.26]}} \\
RECAP (64) & \shortstack[r]{+11.18\\{[+9.95, +12.29]}} & \shortstack[r]{+8.17\\{[+7.11, +9.05]}} \\
RECAP (padded 64) & \shortstack[r]{+11.48\\{[+10.47, +12.49]}} & \shortstack[r]{+8.31\\{[+7.42, +9.26]}} \\
RECAP (native $\leq68$) & \shortstack[r]{+12.91\\{[+11.49, +14.21]}} & \shortstack[r]{+9.48\\{[+8.26, +10.64]}} \\
\addlinespace[3pt]
\multicolumn{3}{@{}l}{\textit{Qwen3-VL-8B}}\\
RACER & \shortstack[r]{+24.70\\{[+23.85, +25.67]}} & \shortstack[r]{+23.30\\{[+22.44, +24.25]}} \\
RECAP (64) & \shortstack[r]{+9.20\\{[+8.39, +10.14]}} & \shortstack[r]{+8.27\\{[+7.47, +9.19]}} \\
RECAP (padded 64) & \shortstack[r]{+10.46\\{[+9.02, +11.81]}} & \shortstack[r]{+9.49\\{[+8.07, +10.80]}} \\
RECAP (native $\leq68$) & \shortstack[r]{+10.12\\{[+9.15, +11.25]}} & \shortstack[r]{+9.19\\{[+8.23, +10.29]}} \\
\bottomrule
\end{tabular*}
\end{table}

P95 request latency decreases from 3.44 to 3.19\,s on Qwen2.5-VL and from 4.96 to 4.52\,s on Qwen3-VL. ReCue has median within-request timing CVs of 1.38\% and 0.94\%, respectively.

\begin{table}[!htbp]
\centering\small
\setlength{\tabcolsep}{2.5pt}
\renewcommand{\arraystretch}{1.03}
\caption{Request-latency percentiles and within-request timing variability across three repeats.}
\label{tab:supp_repeated_variability}
\begin{tabular*}{\columnwidth}{@{\extracolsep{\fill}}lrrrr@{}}
\toprule
Method & P50 (s) & P95 (s) & \shortstack{Median\\CV (\%)} & \shortstack{P90\\CV (\%)} \\
\midrule
\multicolumn{5}{@{}l}{\textit{Qwen2.5-VL-7B}}\\
AR & 2.36 & 8.10 & 0.61 & 1.58 \\
RACER & 1.37 & 3.88 & 1.07 & 2.50 \\
RECAP (64) & 1.22 & 3.44 & 1.46 & 3.62 \\
RECAP (padded 64) & 1.20 & 3.53 & 1.35 & 3.40 \\
RECAP (native $\leq68$) & 1.20 & 3.56 & 1.41 & 3.74 \\
ReCue & 1.13 & 3.19 & 1.38 & 2.81 \\
\addlinespace[3pt]
\multicolumn{5}{@{}l}{\textit{Qwen3-VL-8B}}\\
AR & 9.20 & 9.73 & 1.67 & 4.94 \\
RACER & 4.15 & 5.46 & 0.71 & 1.65 \\
RECAP (64) & 3.60 & 4.96 & 1.42 & 3.32 \\
RECAP (padded 64) & 3.53 & 5.00 & 1.12 & 2.88 \\
RECAP (native $\leq68$) & 3.61 & 4.98 & 1.33 & 3.21 \\
ReCue & 3.29 & 4.52 & 0.94 & 2.05 \\
\bottomrule
\end{tabular*}
\end{table}

\subsection{Resident Resources and Cache Work}
Predictor, bank, and history tensors occupy 48.07\,MiB for Qwen2.5-VL and 51.49\,MiB for Qwen3-VL, with shared storage counted once (Tab.~\ref{tab:supp_resources}). History keys, token-embedding keys, and increments use FP32; counts also include successor IDs and source indices. The table reports preallocated KV capacity and GPU peaks with common loaded assets. Copy rows, continuations, temporary buffers, and runtime allocations contribute additional storage.

\begin{table}[!htbp]
\centering\small
\setlength{\tabcolsep}{2.5pt}
\renewcommand{\arraystretch}{1.03}
\caption{Resident tensor storage and allocated KV capacity.}
\label{tab:supp_resources}
\begin{tabular*}{\columnwidth}{@{\extracolsep{\fill}}lrr@{}}
\toprule
Measurement & Qwen2.5-VL & Qwen3-VL \\
\midrule
Predictor (MiB) & 4.40 & 4.42 \\
Source bank (MiB) & 32.92 & 34.82 \\
History (MiB) & 10.75 & 12.25 \\
KV capacity (MiB) & 224 & 576 \\
Peak GPU (GiB) & 16.07 & 17.27 \\
\bottomrule
\end{tabular*}
\end{table}

\paragraph{State storage and candidate readout.}
History stores state increments in the target's hidden dimension $d$ and state keys for similarity matching. The language head reads the estimated state on the current candidate IDs, including IDs absent from the selected source's top-128 list. For a fixed historical source and a linear head,
\begin{equation}
\begin{aligned}
W(q_t+\Delta_j)
 &= Wq_t+W(h_j^+-h_j^-)\\
 &= Wq_t+Wh_j^+-Wh_j^-.
\end{aligned}
\label{eq:supp_logit_delta}
\end{equation}
This identity also permits a vocabulary-wide score representation. For 256 FP32 records, increments occupy 3.5/4.0\,MiB on Qwen2.5-VL/Qwen3-VL, versus 148.5/148.4\,MiB for vocabulary-wide score increments.

Matching examines at most 256 local records and 16 bank slots; scoring projects at most 256 candidate IDs. Circular history bounds storage as output grows. In Fig.~\ref{fig:recue_cost_benefit}'s MMMU-Pro request, matching, prediction, scoring, and tree construction account for 4.25\% of decode time. History refresh is included in verification and update.

\section{Sanity Checks}
\label{sec:supp_sanity}

\subsection{Task Quality}

ReCue and AR obtain identical task scores in all 15 comparisons in Tab.~\ref{tab:supp_sanity_image_quality}. Each benchmark uses 100 questions, greedy decoding, and natural EOS or a 1,024-token cap. We use normalized exact match for GQA short answers, official soft accuracy for TextVQA short answers, and option accuracy for MMMU-Pro option letters.

\begin{table}[!htbp]
\centering\small
\setlength{\tabcolsep}{2.5pt}
\renewcommand{\arraystretch}{1.03}
\caption{Task scores (\%) under greedy decoding.}
\label{tab:supp_sanity_image_quality}
\begin{tabular*}{\columnwidth}{@{\extracolsep{\fill}}lrrrrrr@{}}
\toprule
& \multicolumn{2}{c}{GQA} & \multicolumn{2}{c}{TextVQA} & \multicolumn{2}{c}{MMMU-Pro} \\
\cmidrule(lr){2-3}\cmidrule(lr){4-5}\cmidrule(lr){6-7}
Backend & AR & ReCue & AR & ReCue & AR & ReCue \\
\midrule
\multicolumn{7}{@{}l}{\textit{Qwen2.5-VL-7B}}\\
Native & 54.0 & 54.0 & 86.5 & 86.5 & 42.0 & 42.0 \\
SGLang & 53.0 & 53.0 & 87.2 & 87.2 & 41.0 & 41.0 \\
\addlinespace[3pt]
\multicolumn{7}{@{}l}{\textit{Qwen3-VL-8B}}\\
Native & 57.0 & 57.0 & 72.8 & 72.8 & 41.0 & 41.0 \\
SGLang & 57.0 & 57.0 & 73.4 & 73.4 & 41.0 & 41.0 \\
\addlinespace[3pt]
\multicolumn{7}{@{}l}{\textit{LLaVA-NeXT Vicuna-7B}}\\
Native & 53.0 & 53.0 & 69.2 & 69.2 & 21.0 & 21.0 \\
\bottomrule
\end{tabular*}
\end{table}

\subsection{Verification and Cache Checks}

For TextVQA short answers on the native backend, ReCue matches greedy AR token for token through EOS on all 100 requests for each Qwen model (Tab.~\ref{tab:supp_sanity_mechanisms}).

All 24,576 tests with an eight-token finite-state target reproduce the serially verified paths and EOS stops at tree budgets of 1, 8, and 64 nodes (Tab.~\ref{tab:supp_sanity_mechanisms}). On 100 requests per Qwen target, source-selection replay (Sec.~\ref{sec:supp_estimate_quality}) keeps prefix KV, candidate pools and their sources, and verifier predictions fixed over 13,684 rounds. ReCue tokens and accepted paths match the native implementation on every request.

\begin{table}[!htbp]
\centering\small
\setlength{\tabcolsep}{3pt}
\renewcommand{\arraystretch}{1.03}
\caption{Verification and output agreement.}
\label{tab:supp_sanity_mechanisms}
\begin{tabular*}{\columnwidth}{@{\extracolsep{\fill}}lr@{}}
\toprule
Check & Passed / total \\
\midrule
Greedy tree verification & 24,576 / 24,576 \\
History transfer and fallback & 300 / 300 \\
History commit and wraparound & 81 / 81 \\
Current-state preservation & 381 / 381 \\
Request-history reset & 12 / 12 \\
Candidate top-4 readout & 36 / 36 \\
Prefix-verifier isolation & 13,684 / 13,684 \\
Native ReCue vs. replay & 200 / 200 \\
\midrule
Greedy AR: Qwen2.5-VL (TextVQA) & 100 / 100 \\
Greedy AR: Qwen3-VL (TextVQA) & 100 / 100 \\
\bottomrule
\end{tabular*}
\end{table}

The cache tests confirm the expected transitions and history resets and verify that the current query remains unchanged until commit. On controlled tensors, candidate readout returns the same top four distinct tokens as FP64 scoring in all 36 pools, with a maximum score error of $2.39\times10^{-5}$.

\FloatBarrier

\end{document}